\documentclass[letterpaper]{article} 
\usepackage{aaai2026}  
\usepackage{times}  
\usepackage{helvet}  
\usepackage{courier}  
\usepackage[hyphens]{url}  
\usepackage{graphicx} 
\usepackage{natbib}  
\usepackage{caption} 
\usepackage{algorithm}
\usepackage[noend]{algorithmic}

\usepackage{newfloat}
\usepackage{listings}
\DeclareCaptionStyle{ruled}{labelfont=normalfont,labelsep=colon,strut=off} 
\floatstyle{ruled}
\newfloat{listing}{tb}{lst}{}
\floatname{listing}{Listing}
\usepackage{booktabs}
\usepackage{amsmath}
\usepackage{amsthm}
\usepackage{mathtools}
\usepackage{amssymb}
\usepackage{enumitem}
\usepackage{subcaption}
\usepackage{xcolor} 

\newtheorem{theorem}{Theorem}
\newtheorem{corollary}{Corollary}
\newtheorem{lemma}[theorem]{Lemma}

\newtheorem{proposition}{Proposition}

\theoremstyle{definition}
\newtheorem{definition}{Definition}

\newtheorem{example}{Example}

\theoremstyle{remark}

\title{Center-Outward q-Dominance: A Sample-Computable Proxy for Strong Stochastic Dominance in Multi-Objective Optimisation}
\author {
    Robin van der Laag\textsuperscript{\rm 1},
    Hao Wang\textsuperscript{\rm 1},
    Thomas B\"ack\textsuperscript{\rm 1},
    Yingjie Fan\textsuperscript{\rm 1}
}
\affiliations {
    \textsuperscript{\rm 1}Leiden Institute for Advanced Computer Science\\
    r.p.van.der.laag@liacs.leidenuniv.nl, h.wang@liacs.leidenuniv.nl, T.H.W.Baeck@liacs.leidenuniv.nl,
    y.fan@liacs.leidenuniv.nl
}

\begin{document}

\maketitle

\begin{abstract}
Stochastic multi-objective optimization (SMOOP) requires ranking multivariate distributions; yet, most empirical studies perform scalarization, which loses information and is unreliable. Based on the optimal transport theory, we introduce the \emph{center-outward $q$-dominance relation} and prove it implies strong first-order stochastic dominance (FSD). Also, we develop an empirical test procedure based on $q$-dominance, and derive an explicit sample size threshold, $n^*(\delta)$, to control the Type I error.
We verify the usefulness of our approach in two scenarios: (1) as a ranking method in hyperparameter tuning; (2) as a selection method in multi-objective optimization algorithms. For the former, we analyze the final stochastic Pareto sets of seven multi-objective hyperparameter tuners on the YAHPO-MO benchmark tasks with $q$-dominance, which allows us to compare these tuners when the expected hypervolume indicator (HVI, the most common performance metric) of the Pareto sets becomes indistinguishable. For the latter, we replace the mean value-based selection in the NSGA-II algorithm with $q$-dominance, which shows a superior convergence rate on noise-augmented ZDT benchmark problems.
These results establish center-outward $q$-dominance as a principled, tractable foundation for seeking truly stochastically dominant solutions for SMOOPs.

\end{abstract}


\section{Introduction}\label{sec:introduction}

Decision-making often involves balancing conflicting objectives under stochasticity. 
Hyperparameter optimization (HPO) is a typical example, where each run of an algorithm produces a vector of random outcomes (validation accuracy, training time, memory footprint, etc.) whose joint distribution reflects noise in the data and the learning procedure itself.
More generally, each candidate solution of a stochastic multi-objective optimization problem (SMOOP) is associated with a distribution of multi-objective performance. Comparing different solutions, therefore, requires ranking these distributions rather than comparing single deterministic points.

Studies on SMOOPs have typically resorted to one of the following pragmatic strategies:
\begin{itemize}
    \item Scalarization: turns the problem into a single-objective stochastic optimization problem by applying some predetermined scalarization function to the objectives \cite{CABALLERO2004633,Abdelaziz2012,DruganN13,trappler2025multiobjective};
    \item Single-sample iterations: solving the optimization problem by evaluating candidate solutions using a single Monte-Carlo draw per iteration \cite{1438403};
    \item Moment surrogates: replacing each random objective by a summary statistic, such as the mean \cite{Fliege2011}, variance, (C)VaR~\cite{DaultonCBOZB22}, or its worst-case scenario, e.g., minimax robustness~\cite{EHRGOTT201417}, multi-objective multi-armed bandits~\cite{LuWHZ19,XuK23}, thereby converting the problem into a deterministic one;
    \item Weak stochastic dominance testing: comparing the empirical cumulative distribution functions (CDF) of the candidate solutions \cite{10.1007/3-540-44719-9_22};
\end{itemize}
While these strategies ease computation, they also discard some parts of the distributional structure and can lead to suboptimal solutions.

A principled alternative to compare distributions is \emph{strong first-order stochastic dominance} (FSD): a distribution $A$ is preferred to $B$ if every non-decreasing utility considers samples from $A$ at least as good as samples from $B$ in expectation. Strong FSD is attractive; it preserves distributional information and imposes no arbitrary scalarization. 
Section~\ref{sec:background} contains the necessary background on (stochastic) multi-objective optimization problems and stochastic dominance.

Testing for strong FSD through samples can be computationally difficult in higher dimensions.
Recent work by \citeauthor{Rioux2025}, \citeyear{Rioux2025} proposes a statistic that assesses multivariate \emph{almost} stochastic dominance by constructing an optimal coupling between two distributions.

In this paper, we take a different route. 
We introduce \emph{center-outward $q$-dominance} (formalized in Section~\ref{sec:COD}): utilizing center-outward distributions and quantile functions \cite{Hallin2017}, which we cover in Section~\ref{sec:CO}, we construct maps from each distribution to the same uniform distribution $U_d$ on the unit ball. 
With this construction, all pairwise comparisons reuse a single common reference frame; we thus only compare samples with identical center-outward rank and sign, ruling out the possibility of incoherent couplings when comparing more than two distributions, such as $X_i	\leftrightarrow Y_j$, $X_i	\leftrightarrow Z_k$, but not $Y_j 	\leftrightarrow Z_k$. Consequently, we only need to compute one coupling per distribution, independent of the number of distributions we wish to compare, in contrast to the method in \cite{Rioux2025}, which requires the computation of a coupling for each order pair of distributions.
We show that if $q$-dominance holds for all quantiles, we obtain strong FSD; if it only holds up to some quantile, we get a natural relaxation.
Because the center-outward ranks are computed once, the entire set of relaxations is obtained at no extra cost. \citeauthor{Rioux2025} can similarly vary their threshold $\varepsilon_0$ for free once an entropic-OT coupling is fixed, but exploring different regularization levels $\lambda$ or functions $h$ still requires a new coupling to be obtained, whereas our method has no tunable hyperparameters.
Furthermore, our finite-sample test admits an explicit minimum sample size threshold $n^*(\delta)$ that guarantees a user-specified Type-I error $\delta$. Lastly, we propose a sorting procedure that enables practical ranking of solutions through pairwise $q$-dominance and its relaxation.

We verify the usefulness of our approach with two complementary experiments in Section~\ref{sec:experiments}. 
Firstly, we re-analyze the final stochastic Pareto sets of seven multi-objective HPO methods on the YAHPO-MO benchmark tasks \cite{pfisterer22a} with $q$-dominance, as opposed to the expected Hypervolume Indicator (HVI). We find that, with $q$-dominance, we can still meaningfully compare the different methods even when the expected HVI of the Pareto sets becomes indistinguishable.
Secondly, we apply $q$-dominance to the well-known NSGA-II algorithm~\cite{nsga2} for multi-objective optimization, where $q$-dominance is used to compare different solutions under uncertainty of the objectives. Compared to the default mean value-based or single-sample-based selection, $q$-dominance enables the algorithm to converge significantly faster on noise-augmented ZDT benchmark problems \cite{ZDT}. 

\section{Background}\label{sec:background}
In this section, we will provide an introduction to the background of multi-objective optimization and stochastic dominance. 
\subsection{Multi-Objective Optimization Problems}\label{sec:background:MOOP} 
A multi-objective optimization problem (MOOP) is typically formulated as
\begin{equation}\label{eq:MOOP}
    \begin{alignedat}{2}
        &\max_x &&\left( f_1(x), \dots, f_m(x) \right)\\
        &\text{subject to}\qquad && x\in\mathcal{X},
    \end{alignedat}
\end{equation}
where $\mathcal{X}\subseteq\mathbb{R}^d$ is the solution space, and $f: \mathcal{X}\to\mathbb{R}^m$,  with $m\geq 2$, the vector-valued objective function.

\begin{definition}\label{def:pareto}
    A feasible solution $x\in\mathcal{X}$ is said to \textit{Pareto dominate} another solution $y\in\mathcal{X}$, if
    \begin{enumerate}
        \item For all $i\in\{1,\dots,m\}$, $f_i(x)\geq f_i(y)$, and
        \item There exists an $i\in\{1,\dots,m\}$, such that $f_i(x) > f_i(y)$.
    \end{enumerate}
    A solution $x^*\in\mathcal{X}$ along with its image point $f(x^*)$ is called \textit{Pareto optimal} if there does not exist another solution that dominates it.
    The set of Pareto optimal points is called the \textit{Pareto front}.
\end{definition}

Assume a probability space $(\Omega, \mathcal{A}, P)$. Stochastic multi-objective optimization problems (SMOOP) arise when the objective function is stochastic, i.e., $f:\mathcal{X}\times\Omega \to \mathbb{R}^m$. With $\omega\in\Omega$, an SMOOP can be written as
\begin{equation}\label{eq:SMOOP}
    \begin{alignedat}{2}
        &\max_x &&\left( f_1(x,\omega), \dots, f_m(x,\omega) \right)\\
        &\text{subject to}\qquad && x\in\mathcal{X},
    \end{alignedat}
\end{equation}

Notably, the Pareto optimal solution to an SMOOP depends on the realization $\omega$.
Two typical approaches to interpret the maximization, and thus the comparison, of random vectors are the \textit{multi-objective method} and the \textit{stochastic method} \cite{Abdelaziz2012}.

The \textit{multi-objective method} defines, for each component of the objective function $f_i$, a vector $( \mathcal{F}_i^{(1)}(f_i(x,\omega)), \dots, \mathcal{F}_i^{(r_i)}(f_i(x,\omega)))$ of one or more statistical functionals $\mathcal{F}_i^{(j)}$ of the random variable $f_i(x,\omega)$, common functionals are the expectation and the variance. With this, we reformulate the SMOOP \eqref{eq:SMOOP} as an MOOP with the $r_1+\dots+r_m$ functionals as the deterministic objectives.

The \textit{stochastic method} scalarizes the random objectives $f_1(x,\omega),\dots,f_m(x,\omega)$ using a function $u:\mathbb{R}^m\to\mathbb{R}$, and reformulates SMOOP \eqref{eq:SMOOP} as a single-objective stochastic optimization problem.

Both of these methods have obvious downsides. The multi-objective method reduces the joint distribution of the objectives to a finite set of summary statistics (the functionals), losing information about the distribution in the process, particularly because the dependency between the random objectives is not taken into account. The stochastic method does not lose this dependency between objectives; however, it does require that the scalarization function is known, and the solution does not generalize to other scalarization functions.

A different approach is to utilize the concept of multivariate stochastic dominance, which allows us to compare random vectors directly without losing any information.

\subsection{Stochastic Dominance}\label{sec:background:SD}
Firstly, let us introduce the concepts of first-order stochastic dominance (FSD) in the scalar case.
\begin{definition}\label{def:scalarFSD}
    Let $X$ and $Y$ be real-valued random variables with cumulative distribution functions $F_X$ and $F_Y$. We say that $X$ first-order stochastically dominates $Y$, written $X \succeq_1 Y$, if and only if any (and hence all) of the following equivalent conditions hold:
    \begin{enumerate}[label=(\roman*)]
        \item $F_X(z) \leq F_Y(z)$ for all $z\in\mathbb{R}$;
        \item $\mathbb{E}[u(X)] \geq \mathbb{E}[u(Y)]$ for all non-decreasing utility functions $u:\mathbb{R}\to\mathbb{R}$.
    \end{enumerate}
\end{definition}

A natural first step is to lift Definition~\ref{def:scalarFSD} to $\mathbb{R}^d$ by replacing the scalar CDF with the joint CDF
\begin{equation*}
    F_\mathbf{X}(\mathbf{z})=\mathbb{P}(\mathbf{X}_1\leq \mathbf{z}_1,\dots,\mathbf{X}_d\leq \mathbf{z}_d)
\end{equation*}
and by letting the test set be all non-decreasing utility functions $u:\mathbb{R}^d\to\mathbb{R}$.
Crucially, in dimension $d>1$, these two characterizations diverge, giving rise to the weak (via CDFs) and strong (via utility functions) notions of multivariate FSD.
\begin{definition}[Weak FSD]\label{def:weakFSD}
    Let $\mathbf{X}, \mathbf{Y}\in\mathbb{R}^d$ be real-valued random vectors with joint CDFs $F_\mathbf{X}$ and $F_\mathbf{Y}$. We say that $\mathbf{X}$ weakly stochastically dominates $\mathbf{Y}$, written $\mathbf{X}\succeq_w \mathbf{Y}$, if and only if
    \begin{equation*}
        F_\mathbf{X}(\mathbf{z}) \leq F_\mathbf{Y}(\mathbf{z}), \qquad \forall \mathbf{z}\in\mathbb{R}^d.
    \end{equation*}
\end{definition}

\begin{definition}[Strong FSD]\label{def:strongFSD}
    We say that $\mathbf{X}$ strongly stochastically dominates $\mathbf{Y}$, written $\mathbf{X}\succeq_1\mathbf{Y}$, if and only if
    \begin{equation*}
        \mathbb{E}[u(\mathbf{X})] \geq \mathbb{E}[u(\mathbf{Y})],
    \end{equation*}
    for all non-decreasing utility functions $u:\mathbb{R}^d\to\mathbb{R}$.
\end{definition}

To relate the two orders, we first recall an equivalent characterization of strong FSD from \cite{Sriboonchita2009} that replaces utilities by probabilities on so-called upper sets.

\begin{proposition}\label{prop:strongfsd-upperset}
    For random vectors $\mathbf{X}, \mathbf{Y}\in\mathbb{R}^d$ the following are equivalent:
    \begin{enumerate}[label=(\roman*)]
        \item $\mathbf{X} \succeq_1 \mathbf{Y}$ (Definition~\ref{def:strongFSD}).
        \item For every upper set $M\subseteq\mathbb{R}^d$,
        \begin{equation*}
            \mathbb{P}(\mathbf{X} \in M) \geq \mathbb{P}(\mathbf{Y}\in M).
        \end{equation*}
        Here an upper set is any $M\subseteq\mathbb{R}^d$ such that for any $\mathbf{z},\mathbf{w}\in\mathbb{R}^d$ with $\mathbf{w}\geq\mathbf{z}$ we have that $\mathbf{w}\in M$ whenever $\mathbf{z} \in M$.
    \end{enumerate}
\end{proposition}
\begin{proof}
    We refer to \cite{Sriboonchita2009}.
\end{proof}

The upper set view makes the hierarchy between the two orders apparent, yielding the following result.

\begin{theorem}
    For every dimension $d\geq 1$ and every pair of random vectors $\mathbf{X},\mathbf{Y}\in\mathbb{R}^d$,
    \begin{equation*}
        \mathbf{X}\succeq_1 \mathbf{Y} \implies \mathbf{X}\succeq_w \mathbf{Y}.
    \end{equation*}
\end{theorem}
\begin{proof}
    For $\mathbf{m}=(m_1,\dots,m_d)\in\mathbb{R}^d$ we have that $1-F_\mathbf{X}(\mathbf{m})=\mathbb{P}(\mathbf{X}\geq\mathbf{m})=\mathbb{P}(\mathbf{X}\in [m_1,\infty)\times\dots\times [m_d,\infty))$ and similarly for $\mathbf{Y}$. The set $[m_1,\infty)\times\dots\times [m_d,\infty)$ is an upper set in $\mathbb{R}^d$, thus if $\mathbb{P}(\mathbf{X}\in M)\geq \mathbb{P}(\mathbf{Y}\in M)$ for all upper sets $M\in\mathcal{M}$, then it also holds for the special form above. This gives us that $1-F(\mathbf{m})\geq 1-G(\mathbf{m})$, implying $\mathbf{X}\succeq_w \mathbf{Y}$.
\end{proof}
The converse does not hold for dimensions $d>1$, as the next counterexample from \cite{Kopa2018StrongAW} demonstrates.
\begin{example}\label{example:weak<strong}
    Consider the two random vectors
    \begin{align*}
        \mathbf{X}=\begin{cases}
            (0,1),\quad \text{w.p. }1/2,\\
            (1,0),\quad \text{w.p. }1/2.
        \end{cases}\hfill
        \mathbf{Y}=\begin{cases}
            (0,0),\quad \text{w.p. }1/2,\\
            (1,1),\quad \text{w.p. }1/2.
        \end{cases}
    \end{align*}
    Then $F_\mathbf{X}(\mathbf{z})\leq F_\mathbf{Y}(\mathbf{z})$ for all $\mathbf{z}\in\mathbb{R}^2$ and thus $\mathbf{X}\succeq_w \mathbf{Y}$. Consider the upper set $M=\{\mathbf{m}\in\mathbb{R}^2 : m_1+m_2\geq\frac{3}{2}\}$, then $0=\mathbb{P}(\mathbf{X}\in M) < \mathbb{P}(\mathbf{Y}\in M) = \frac{1}{2}$, and thus $\mathbf{X}\not\succeq_{1}\mathbf{Y}$.
\end{example}
Example~\ref{example:weak<strong} makes the dilemma explicit: in dimensions $d>1$ weak FSD can be too permissive, declaring $\mathbf{X}\succeq_w\mathbf{Y}$ even though strong FSD rejects that claim, i.e. $\mathbf{X}\not\succeq_1\mathbf{Y}$.
Equivalently, there exists a non-decreasing utility $u$---for instance $u(x_1,x_2)=\exp(x_1+x_2)$---for which $\mathbb{E}[u(\mathbf{X})] < \mathbb{E}[u(\mathbf{Y})]$. 

As noted in the introduction, testing for strong FSD can be computationally difficult without full knowledge of the joint distribution. \citeauthor{Rioux2025} use the following theorem from \cite{stochasticorder-book} for their optimal transport approach.
\begin{theorem}\label{thm:FSD-couplingequiv}
    The random vectors $\mathbf{X},\mathbf{Y}$ satisfy $\mathbf{X}\succeq_1\mathbf{Y}$, if and only if there exists a coupling $(\widehat{\mathbf{X}},\widehat{\mathbf{Y}})$ of $(\mathbf{X},\mathbf{Y})$ satisfying $\mathbb{P}(\widehat{\mathbf{X}}\geq \widehat{\mathbf{Y}})=1$.
\end{theorem}
\citeauthor{Rioux2025} then provides the following lemma, casting this theorem into the context of optimal transport.
\begin{lemma}
    Let $P_\mathbf{X},P_\mathbf{Y}$ denote the distributions of the random vectors $\mathbf{X}$ and $\mathbf{Y}$, respectively. 
    Then $\mathbf{X}\succeq_1\mathbf{Y}$ if 
    \begin{equation*}
        \inf_{\pi\in\Pi(P_\mathbf{X},P_\mathbf{Y})} \int c\ \mathrm{d}\pi = 0,
    \end{equation*}
    where $c:\mathbb{R}^d\times\mathbb{R}^d\to\mathbb{R}_+$ is the cost function $c(x,y)=\mathbf{1}_{\{x\leq y\}}$ and $\Pi(P_\mathbf{X},P_\mathbf{Y})$ denotes the set of all couplings of $(P_\mathbf{X},P_\mathbf{Y})$.
\end{lemma}

When more than two distributions $P_1, P_2,\dots$ must be compared, this method requires solving an optimal transport problem for every ordered pair $(P_i, P_j)$. Because each coupling is optimized in isolation a sample $\mathbf{X}\sim P_1$ that is matched to $\mathbf{Y}\sim P_2$ under $\pi_{1,2}$ and to $\mathbf{Z}\sim P_3$ under $\pi_{1,3}$ need not have $\mathbf{Y}$ matched to $\mathbf{Z}$ under $\pi_{2,3}$. The resulting family of couplings, therefore, lacks a common reference frame and is difficult to interpret jointly. 

In the next section, we introduce \emph{center-outward ranks and signs}, providing the missing common reference frame that underpins our definition of $q$-dominance.

\section{Center-Outward Ranks and Signs}\label{sec:CO}
In this section, we define the center-outward distribution and quantile function, which are rooted in the main result of \cite{McCann1995} and further developed in \cite{Hallin2017} and \cite{Hallin2021}.

Throughout this section, we make use of the following notation. 
Let $\mu_d$ denote the Lebesgue measure over $\mathbb{R}^d$ equipped with its Borel $\sigma$-field $\mathcal{B}_d$. Furthermore, denote by $\mathcal{P}_d$ the family of Lebesgue-absolutely continuous distributions over $(\mathbb{R}^d, \mathcal{B}_d)$. We use the notation $T\#P_1=P_2$ for the distribution $P_2$ of $T(X)$, where $X\sim P_1$, and say that $T$ is pushing $P_1$ forward to $P_2$.
Lastly, let $\mathcal{S}_{d-1}$, $\mathbb{S}_d$, and $\overline{\mathbb{S}}_d$ denote the unit sphere, the open unit ball, and the closed unit ball in $\mathbb{R}^d$, respectively.

\subsection{Theoretical Definition}\label{sec:CO:def}
We restate the main result of \cite{McCann1995}.
\begin{theorem}[\citeauthor{McCann1995} \citeyear{McCann1995}]
For two distributions $P_1,P_2 \in\mathcal{P}_d$ the following statements hold:
\begin{enumerate}[label=(\roman*)]
    \item the class of functions
    \begin{equation*}
        \nabla\Psi_{P_1;P_2} \coloneq \{ \nabla\psi\ \vert\ \psi:\mathbb{R}^d\to\mathbb{R},\ \nabla\psi\#P_1 = P_2 \},
    \end{equation*}
    where $\psi$ is convex and lower semi-continuous, is non-empty;
    \item if $\nabla\psi',\nabla\psi'' \in \nabla\Psi_{P_1;P_2}$, they coincide $P_1$-a.s., that is $P_1\left(\{\mathbf{x}\ \vert\ \nabla\psi'(\mathbf{x}) \not= \nabla\psi''(\mathbf{x})\}\right)=0$;
    \item if $P_1$ and $P_2$ have finite moments of order two, any element of $\nabla\Psi_{P_1;P_2}$ is an optimal quadratic transport pushing $P_1$ forward to $P_2$, i.e. it is a solution to
    \begin{equation*}
        \inf_{\gamma\in\Pi(P_1,P_2)} \int_{\mathbb{R}^d\times\mathbb{R}^d}\lVert x-y\rVert^2\ \mathrm{d}\gamma(x,y),
    \end{equation*}
    where $\Pi(P_1,P_2)$ denotes the set of joint distributions with marginals $P_1$ and $P_2$.
\end{enumerate}    
\end{theorem}

Denote by $U_d$ the uniform distribution over $\mathbb{S}_d$, which is the product of the uniform over the unit sphere and the uniform over the unit interval. 
The center-outward distribution and quantile functions are then defined as follows.
\begin{definition}
    The center-outward quantile function $\mathbf{Q}^\pm$ of $P\in\mathcal{P}_d$ is the a.e. unique element $\nabla\psi\in\nabla\Psi_{U_d;P}$, such that $\psi$ satisfies 
    \begin{equation*}
        \psi(\mathbf{u}) = \infty, \text{ for }\lVert\mathbf{u}\rVert>1
    \end{equation*}
    \noindent and
    \begin{equation*}
        \psi(\mathbf{u})=\liminf_{\mathbb{S}_d\ni \mathbf{v}\to\mathbf{u}} \psi(\mathbf{v}),\text{ for }\lVert\mathbf{u}\rVert=1.
    \end{equation*}
\end{definition}
\begin{definition}
    Call $\mathbf{F}^\pm \coloneqq \nabla\phi$ the center-outward distribution function, where $\phi$ is defined as the Legendre transform
    \begin{equation*}
        \phi(\mathbf{x}) \coloneqq \psi^*(\mathbf{x}) \coloneqq \sup_{\mathbf{u}\in\mathbb{S}_d}(\langle\mathbf{u},\mathbf{x}\rangle - \psi(\mathbf{u})), \quad \mathbf{x}\in\mathbb{R}^d.
    \end{equation*}
\end{definition}
The following propositions from \cite{Hallin2021} summarize the main properties of these functions.
\begin{proposition}
    Let $\mathbf{Z}\sim P\in\mathcal{P}_d$ and let $\mathbf{F}^\pm$ be the center-outward distribution function of $P$, then
    \begin{enumerate}[label=(\roman*)]
        \item $\mathbf{F}^\pm$ takes values in $\overline{\mathbb{S}}_d$ and $\mathbf{F}^\pm \# P = U_d$. Thus $\mathbf{F}^\pm$ is a probability-integral transformation;
        \item $\lVert\mathbf{F}^\pm(\mathbf{Z})\rVert$ is uniform over $[0,1]$,
        $\mathbf{S}(\mathbf{Z})={\mathbf{F}^\pm(\mathbf{Z})/\lVert\mathbf{F}_\pm(\mathbf{Z})\rVert}$ is uniform over $\mathcal{S}_{d-1}$, and they are mutually independent;
        \item $\mathbf{F}^\pm$ entirely characterizes $P$;
        \item for $d=1$, $\mathbf{F}^\pm$ coincides with $2F-1$, where $F$ is the tradition distribution function.
    \end{enumerate}
\end{proposition}

\begin{definition}\label{def:q-region}
    For $q\in(0,1)$ we call
    \begin{equation*}
        \mathcal{C}(q)\coloneqq \mathbf{Q}^\pm(q\mathcal{S}_{d-1})=\{\mathbf{z}\in\mathbb{R}^d\ \vert\ \lVert\mathbf{F}^\pm(\mathbf{z})\rVert = q\}
    \end{equation*}
    the center-outward quantile counter, and
    \begin{equation*}
        \mathbb{C}(q)\coloneqq \mathbf{Q}^\pm(q\overline{\mathbb{S}}_d)=\{\mathbf{z}\in\mathbb{R}^d\ \vert\ \lVert\mathbf{F}^\pm(\mathbf{z})\rVert \leq q\}
    \end{equation*}
    the center-outward quantile region of order $q$. Furthermore, 
    \begin{equation*}
        \mathbb{C}(0)\coloneqq \bigcap_{0<q<1} \mathbb{C}(q) = \partial\psi(\mathbf{0}).
    \end{equation*}
\end{definition}

\begin{proposition}
    Let $P\in\mathcal{P}_d$ have center-outward quantile function $\mathbf{Q}_\pm$, then
    \begin{enumerate}[label=(\roman*)]
        \item $\mathbf{Q}^\pm\# U_d = P$, and hence $\mathbf{Q}^\pm$ entirely characterizes $P$;
        \item the center-outward quantile region $\mathbb{C}(q)$, for $0<q<1$, has $P$-probability content $q$;
    \end{enumerate}
\end{proposition}

\subsection{Empirical Estimation}\label{sec:CO:emp}
We now examine how to construct empirical counterparts to $\mathbf{F}^\pm$ and $\mathbf{Q}^\pm$ using samples.

Denote by $\mathbf{Z}^{(n)} \coloneqq \left(\mathbf{Z}_1,\dots, \mathbf{Z}_n \right)$ an $n$-tuple of i.i.d. random vectors from distribution $P\in\mathcal{P}$, with density $f$ and center-outward distribution function $\mathbf{F}^\pm$.
For the empirical center-outward distribution function $\widehat{\mathbf{F}}^\pm$ \cite{Hallin2021} propose the following construction.

Assume that $d\geq 2$, and factorize $n=n_R n_S + n_0$ with $n_R,n_S,n_0 \in \mathbb{N}$ and $0\leq n_0 < \min\{n_R,n_S\}$, such that $n_R\to\infty$ and $n_S\to\infty$ as $n\to\infty$. Next, we define a sequence of regular grids over the unit ball $\mathbb{S}_d$ as the intersection between
\begin{itemize}
    \item a regular $n_S$-tuple $\mathfrak{S}^{(n_s)}\coloneqq(\mathbf{u}_1,\dots,\mathbf{u}_{n_S})$ of unit vectors, and
    \item $n_R$ hyperspheres centered at $\mathbf{0}$, with radii $\left\{\frac{1}{n_R+1},\dots,\frac{n_R}{n_R+1}\right\}$,
\end{itemize}
along with $n_0$ copies of the origin. 
The discrete distribution with probability masses $1/n$ at each of the $n_Rn_S$ grid points and probability mass $n_0/n$ at the origin converges weakly to the uniform $U_d$ over the ball $\mathbb{S}_d$. We refer to this grid as the \emph{augmented grid}.

\begin{definition}
    The empirical center-outward distribution function is any mapping
    $\widehat{\mathbf{F}}^\pm: \left(\mathbf{Z}_1,\dots, \mathbf{Z}_n \right) \mapsto \left(\widehat{\mathbf{F}}^\pm(\mathbf{Z}_1),\dots, \widehat{\mathbf{F}}^\pm(\mathbf{Z}_n) \right)$ satisfying
    \begin{equation}\label{eq:def-empcofunc}
        \sum_{i=1}^n \lVert \mathbf{Z}_i - \widehat{\mathbf{F}}^\pm(\mathbf{Z}_i)\rVert^2
        = \min_\pi \sum_{i=1}^n \lVert \mathbf{Z}_{\pi(i)} - \widehat{\mathbf{F}}^\pm(\mathbf{Z}_i)\rVert^2,
    \end{equation}
    where the set $\{ \widehat{\mathbf{F}}^\pm(\mathbf{Z}_i)\ \vert\  i=1,\dots,n \}$ consists of the $n$ points of the augmented grid and $\pi$ ranges over the $n!$ permutations of $\{1,2,\dots,n\}$.
\end{definition}
The assignment problem in Equation \eqref{eq:def-empcofunc} can be solved by, for example, the Hungarian algorithm in $\mathcal{O}(n^3)$.

Along with the definition of the empirical center-outward distribution function $\widehat{\mathbf{F}}^\pm$, we define the following concepts:
\begin{itemize}
    \item center-outward ranks $\widehat{R}_{i} \coloneqq (n_R+1)\lVert\widehat{\mathbf{F}}^\pm(\mathbf{Z}_i)\rVert$;
    \item empirical center-outward quantile contours $\widehat{\mathcal{C}}(\frac{j}{n_R+1}) \coloneqq \{ \mathbf{Z}_i\ \vert\ \widehat{R}_{i}=j\}$ and regions $\widehat{\mathbb{C}}(\frac{j}{n_R+1}) \coloneqq \{ \mathbf{Z}_i\ \vert\ \widehat{R}_{i}\leq j\}$, where $j/(n_R+1)$, $j\in\{0,\dots,n_R\}$, are empirical probability contents, to be interpreted as a quantile order;
    \item center-outward signs: ${\widehat{\mathbf{S}}_i \coloneqq\mathbf{1}_{\left\{ \widehat{\mathbf{F}}^\pm(\mathbf{Z}_i) \not= \mathbf{0} \right\}}  \frac{ \widehat{\mathbf{F}}^\pm(\mathbf{Z}_i)}{\lVert\widehat{\mathbf{F}}^\pm(\mathbf{Z}_i)\rVert}}$, and sign curves $\{\mathbf{Z}_i\ \vert\ \widehat{\mathbf{S}}_i=\mathbf{u}\}$, for ${\mathbf{u}\in\mathfrak{S}^{(n_S)}}$.
\end{itemize}

\begin{figure}[t]
    \centering
    \includegraphics[width=\linewidth]{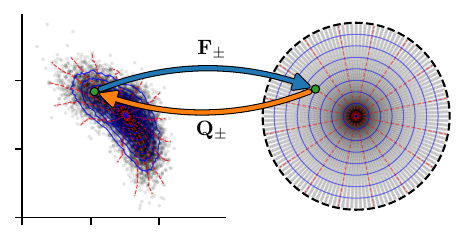}
    \caption{Samples of a bi-variate distribution (left) and the points on the augmented grid (right). Selected center-outward quantile contours are shown in blue and signs in red.}
    \label{fig:co-example}
\end{figure}

In Figure~\ref{fig:co-example}, samples from a bi-variate distribution along with selected quantile contours and signs in the sample space $\mathbb{R}^2$ (left) and in the unit ball $\mathbb{S}_2$ (right) are shown, along with the mappings $\mathbf{F}_\pm$ and $\mathbf{Q}_\pm$.

\section{Center-Outward Dominance Relation}\label{sec:COD}
We now introduce our main contribution: \emph{center-outward $q$-dominance}, a dominance relation between two distributions using their center-outward distribution and quantile functions.

Let $P_1, P_2 \in \mathcal{P}_d$ be two probability distributions with center-outward distribution and quantile functions $\mathbf{F}^\pm_1, \mathbf{Q}^\pm_1$ and $\mathbf{F}^\pm_2, \mathbf{Q}^\pm_2$.
\begin{definition}[$q$-dominance]\label{def:qdom}
    For $q\in[0,1)$ we say that $P_1$ dominates $P_2$ at quantile $q$, writing $P_1\succeq_q P_2$, if for every $\mathbf{y}\in\mathbb{C}_{P_2}(q)$ we have that
    \begin{equation}\label{eq:defqdom}
        \mathbf{x} \coloneqq \mathbf{Q}^\pm_1\left(\mathbf{F}^\pm_2(\mathbf{y})\right) \geq \mathbf{y}.
    \end{equation}
    When $\mathbf{F}^\pm_2(\mathbf{y})=0$ (meaning $\mathbf{y}\in\mathbb{C}_{P_2}(0)$), we interpret ${\mathbf{Q}^\pm_1\left(0\right) \geq \mathbf{y}}$ as ``every $\mathbf{x}\in\mathbb{C}_{P_1}(0)$ satisfies $\mathbf{x}\geq \mathbf{y}$.''
\end{definition}

\subsection{Theoretical Properties}\label{sec:COD:prop}

Definition~\ref{def:qdom} yields three immediate consequences, which we collect below as Corollary~\ref{cor:qdom}, and is connected to strong FSD, as we will show in Theorem~\ref{thm:q->fsd}. 
\begin{corollary}\label{cor:qdom}
    If $P_1$ dominates $P_2$ for a quantile $q\in[0,1)$; $P_1 \succeq_q P_2$, then:
    \begin{enumerate}[label=(\roman*)]
        \item For any $q' \leq q$ we also have that $P_1 \succeq_{q'}P_2$ , since $\mathbb{C}_{P_i}(q')\subseteq\mathbb{C}_{P_i}(q)$.
        \item For any non-decreasing utility $u:\mathbb{R}^d\to\mathbb{R}$,
        \begin{equation*}
            \mathbb{E}[u(\mathbf{X})\ \vert\ \mathbf{X}\in\mathbb{C}_{P_1}(q)] \geq \mathbb{E}[u(\mathbf{Y})\ \vert\ \mathbf{Y}\in\mathbb{C}_{P_2}(q)].
        \end{equation*}
        If $u$ is bounded from above and below by constants $M$ and $m$ respectively, then
        \begin{equation*}
            \mathbb{E}[u(\mathbf{X})] - \mathbb{E}[u(\mathbf{Y})] \geq \Delta_q - (1-q)(M-m),
        \end{equation*}
        where $\Delta_q \coloneqq \int_{\lVert v\rVert \leq q} u(\mathbf{Q}^\pm_1(v)) - u(\mathbf{Q}^\pm_2(v))\ \mathrm{d}v \geq 0$.
        \item There exists a coupling $(\widehat{\mathbf{X}},\widehat{\mathbf{Y}})$ with marginals $P_1,P_2$ such that
        \begin{equation*}
            \mathbb{P}(\widehat{\mathbf{X}}\geq\widehat{\mathbf{Y}})\geq q.
        \end{equation*}
        The bound is attained by the monotone coupling ${\widehat{\mathbf{X}}=\mathbf{Q}^\pm_1(\mathbf{Z})}$ and $\widehat{\mathbf{Y}}=\mathbf{Q}^\pm_2(\mathbf{Z})$, with $\mathbf{Z}\sim U_d$.
    \end{enumerate} 
\end{corollary}

\begin{theorem}\label{thm:q->fsd}
    If $P_1\succeq_q P_2$ for all $q\in[0,1)$ then $P_1$ strongly (and thus also weakly) stochastically dominates $P_2$, i.e. $P_1 \succeq P_2$.
\end{theorem}
\begin{proof}
    By Corollary~\ref{cor:qdom}(iii), the map $\mathbf{Z} \mapsto \left(\mathbf{Q}^\pm_1(\mathbf{Z}), \mathbf{Q}^\pm_2(\mathbf{Z})\right)$ provides a coupling with $\mathbb{P}(\mathbf{X}\geq\mathbf{Y}) = 1$, since $P_1 \succeq_q P_2$ for all $q\in[0,1)$, and Theorem~\ref{thm:FSD-couplingequiv} then yields $P_1 \succeq_1 P_2$.
\end{proof}
Thus, $q$-dominance can be seen as a natural relaxation of strong FSD, where we obtain strong FSD when $q$-dominance holds for all quantiles.

\subsection{Empirical Test}\label{sec:COD:emp}
To transform the theory into a finite-sample decision rule, we discretize \eqref{eq:defqdom} on the augmented grid defined in Section~\ref{sec:CO:emp}.
Let $\mathbf{X}^{(n)}=\left(\mathbf{X}_1,\dots,\mathbf{X}_n\right)$ and $\mathbf{Y}^{(n)}=\left(\mathbf{Y}_1,\dots,\mathbf{Y}_n\right)$ be $n$ i.i.d. samples from $P_1$ and $P_2$ respectively, with empirical center-outward distribution and quantile functions $\widehat{\mathbf{F}}^\pm_1, \widehat{\mathbf{Q}}^\pm_1$ and $\widehat{\mathbf{F}}^\pm_2, \widehat{\mathbf{Q}}^\pm_2$.

\begin{definition}\label{def:emp-qdom}
    For $q \in \left\{0, \frac{1}{n_R+1}, \dots, \frac{n_R}{n_R+1} \right\}$, we have that $\mathbf{X}^{(n)} \succeq_q \mathbf{Y}^{(n)}$ if
    \begin{equation*}
        \widehat{\mathbf{Q}}^\pm_1(q'\mathbf{u}) \geq \widehat{\mathbf{Q}}^\pm_2(q'\mathbf{u}),
    \end{equation*}
    for all $q' \in \left\{0, \frac{1}{n_R+1}, \dots, \frac{\lfloor q(n_R+1)\rfloor}{n_R+1} \right\}$ and all $\mathbf{u}\in\mathfrak{S}^{(n_S)}$. When $n_0\geq 2$ the terms $\widehat{\mathbf{Q}}^\pm_1(0)$ and $\widehat{\mathbf{Q}}^\pm_2(0)$ are multi-valued, in this case we interpret the defining inequality as ``for every $\mathbf{x}\in\widehat{\mathbf{Q}}^\pm_1(0)$ and $\mathbf{y}\in\widehat{\mathbf{Q}}^\pm_2(0)$, we have that $\mathbf{x}\geq\mathbf{y}$.''
\end{definition}

In the following theorem, we prove that, once the sample size is large enough, theoretical $q$-dominance of the distributions $P_1 \succeq_q P_2$ carries over to the empirical maps with high probability.
\begin{theorem}\label{thm:pacbound}
    Let distributions $P_1$ and $P_2$ have center-outward quantile functions $\mathbf{Q}^\pm_1, \mathbf{Q}^\pm_2$ that are bi-Lipschitz continuous with constants $L_1$ and $L_2$.
    Choose $n_R=n^\theta$ and $n_S=n^{1-\theta}$, with $\theta\in \left( \frac{1}{2d}, \frac{d+1}{2d} \right)$ for $d \leq 4$ and $\theta\in \left( \frac{d-2}{d^2}, \frac{2d-3}{d^2} \right)$ for $d\geq 5$.
    If we assume that $P_1\succeq_q P_2$ for every $q\in[0,1)$, then for every confidence level $0<\delta<1$ there exists an explicit threshold $n^*(\delta)$ such that for all $n \geq n^*(\delta)$
    \begin{equation*}
        \mathbf{X}^{(n)} \succeq_q \mathbf{Y}^{(n)},\quad \text{for every } q\in\left\{\frac{j}{n_R+1}\right\}_{j=0}^{n_R},
    \end{equation*}
    with probability at least $1-\delta$.
\end{theorem}
A complete proof, including an explicit formula for $n^*(\delta)$, is provided in Appendix~\ref{apx:proof-pacthm}.

Lastly, we provide a straightforward procedure, outlined in Appendix~\ref{apx:alg} (Algorithm~\ref{alg:q-sort}), that ranks a collection of empirical distributions by iteratively applying center-outward $q$-dominance on a shared augmented grid.
We first build non-dominated fronts for successively smaller $q$ values, and then, within each front, order the remaining distributions by a measure of how close they are to being dominated.

\section{Experiments}\label{sec:experiments}
To demonstrate the usefulness of $q$-dominance, we conduct two numerical studies: (1) to help improve the robust ranking of optimizers in multi-objective hyperparameter optimization tasks; (2) to assist the selection in multi-objective evolutionary algorithms (MOEAs) under noise. 

\subsection{Multi-Objective Hyperparameter Optimization (HPO) Rankings}\label{sec:experiments:HPO}
\citeauthor{pfisterer22a} compare seven optimizers: Random Search, Random Search 4x (Random Search with quadrupled budget), ParEGO \cite{ParEGO}, SMS-EGO \cite{SMS-EGO}, EHVI \cite{EHVI}, MEGO \cite{MEGO}, and MIES \cite{MIES} on 25 different multi-objective hyperparameter optimization problem instances with 2 to 4 objectives. 
They compare these optimizers by computing the normalized Hypervolume Indicator (HVI) of their found Pareto fronts after specific fractions of their budget have been used.
For our approach, we sample $ k \leq 5$ points uniformly at random from the same Pareto fronts at each replication, resulting in $30k$ samples from the optimizer's underlying stochastic Pareto set for each problem instance and at each considered fractional budget.
With these samples, we then rank the optimizers using our $q$-dominance relation, by computing their empirical center-outward quantile maps and sorting them using Algorithm~\ref{alg:q-sort}. We repeat this procedure $100$ times to minimize the influence of the random sampling.
More complete experimental details can be found in Appendix~\ref{apx:exp-hpo}.

\begin{figure}[t]
    \begin{subfigure}{\linewidth}
        \centering
        \includegraphics[width=\linewidth]{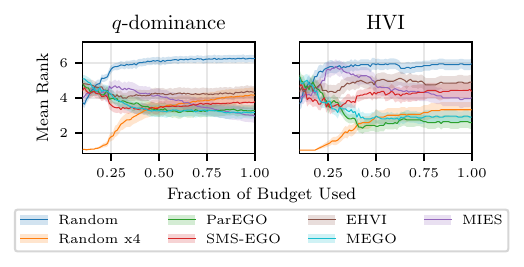}
        \caption{Mean-rank trajectories ($\pm 1$ s.d.).}
        \label{fig:hpo-meanrank}
    \end{subfigure}\\
    \begin{subfigure}{\linewidth}
        \centering
        \includegraphics[width=\linewidth]{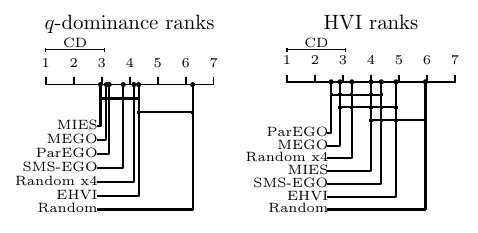}
        \caption{Critical difference diagrams for mean ranks after $100\%$ of budget used.}
        \label{fig:hpo-cd100}
    \end{subfigure}
    \caption{Results of YAHPO-MO benchmarks based on $q$-dominance (left) and HVI (right).}
\end{figure}

\begin{figure}
    \centering
    \includegraphics[width=\linewidth]{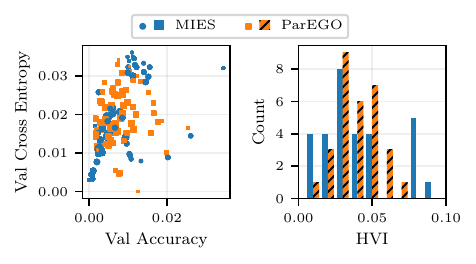}
    \caption{Left: pooled samples from the final stochastic Pareto sets for MIES and ParEGO on \emph{lcbench 167152} (marker size $\propto \text{quantile}^{-1}$).
    Right: histogram of the HVI values for those same Pareto sets.}
    \label{fig:hpo-example}
\end{figure}

In Figure~\ref{fig:hpo-meanrank}, we show the mean rank, based on $q$-dominance (left) and HVI (right), of the different HPO methods as a function of the fraction of budget used. Figure~\ref{fig:hpo-cd100} shows the critical difference plot, with ${\alpha=0.05}$, after $100\%$ of the budget has been used. 

The first difference we note is the spread of the mean ranks between the optimizers after around $50\%$ of the budget has been used. In Figure~\ref{fig:hpo-meanrank} (left), we see that the mean ranks, based on $q$-dominance, of all the methods aside from Random Search are bunched quite close together, and in Figure~\ref{fig:hpo-cd100} (left), we see that none of these methods significantly outperform the others. Whilst for the mean ranks based on HVI, we have a much larger spread on the right side of Figure~\ref{fig:hpo-meanrank} and in Figure~\ref{fig:hpo-cd100}.

Furthermore, in Figure~\ref{fig:hpo-cd100} we see that, based on the $q$-dominance ranking, only EHVI does not significantly improve on Random Search after the entire budget has been used. Contrastingly, based on the HVI ranking, in addition to EHVI, SMS-EGO and MIES also do not significantly improve on Random Search—--a notable difference, specifically for MIES, which on average performs the best according to $q$-dominance.

To further investigate this difference in rankings, we examine the final Pareto sets of MIES and ParEGO, the optimizers that, on average, perform best based on $q$-dominance and HVI, respectively, on the problem instance lcbench 167152 (See Table~\ref{tab:apx-hpo-problems} in Appendix~\ref{apx:exp-hpo}). On this problem, MIES is preferred over ParEGO based on $q$-dominance, with a rank of $1.97 \pm 0.26$ for MIES versus $3.11 \pm 0.60$ for ParEGO, yet ParEGO is preferred over MIES based on HVI, where ParEGO has an HVI of $0.0399 \pm 0.0025$ and MIES has an HVI of $0.0400 \pm 0.0045$ across the 30 replications.
In Figure~\ref{fig:hpo-example}, we show samples from the final Pareto sets of MIES (blue circles) and ParEGO (orange squares) in the left plot, and a histogram of the HVI values of both optimizers over the 30 replications in the right plot.
From these figures, we can infer that MIES has a high probability of achieving better results than ParEGO, shown by the blue, circular points of MIES in the bottom left of the left plot in Figure~\ref{fig:hpo-example} and the left-most bars in the right plot. However, this comes at the cost of slightly higher risk, which we can see from the points of MIES in the upper middle part of the left plot in Figure~\ref{fig:hpo-example} and the right-most bars in the histogram on the right.
By taking the expected value of the HVI, we lose this information, resulting in the HVI ranking favoring ParEGO over MIES, as ParEGO's mean HVI value is slightly lower.

\subsection{Noise Augmented ZDT}

\begin{figure}[t]
    \centering
    \includegraphics[width=\linewidth]{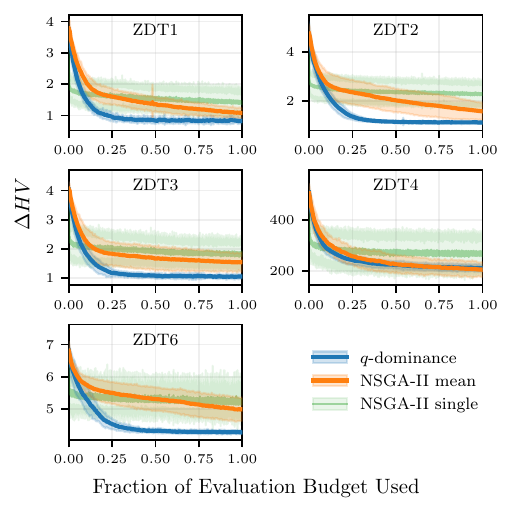}
    \caption{The difference in HV ($\Delta HV$) between the deterministic Pareto front and the expected HV of the solutions at a specific budget used. ZDT5 is excluded as it is a Boolean optimization problem.}
    \label{fig:zdt-averageHV}
\end{figure}

To assess the value of $q$-dominance when directly applied to SMOOPs, we augment the commonly used ZDT benchmark problems \cite{ZDT} with noise in the inputs. Specifically, if $f$ is the objective function, then we consider the optimization problem: $\min_{\mathbf{x}}f(z_1,\dots,z_k)$, where $z_i \sim \mathcal{N}_{[a_i,b_i]}(x_i,\sigma^2)$. Here $a_i,b_i$ indicate the lower- and upper-bound of the decision variable $x_i$, such that $x_i\in[a_i,b_i]$, and $\mathcal{N}_{[a_i,b_i]}(x_i,\sigma^2)$ denotes the truncated normal distribution, with mean $x_i$, variance $\sigma^2$, and truncated to lie within the interval $[a_i,b_i]$.
With this construction, our objective functions become stochastic, and the underlying deterministic functions are never evaluated outside of their domains. 

We consider two simple baselines: NSGA-II with mean-value based selection (denoted by \texttt{NSGA-II mean}), i.e., for each candidate solution we evaluate the objective function $n$ times and compute the mean, and NSGA-II with single sample based selection (\texttt{NSGA-II single}), where we only evaluate the objective function once \cite{nsga2}.
For our method, we replace the selection with the $q$-dominance sorting procedure (Algorithm~\ref{alg:q-sort}), computed on $n$ objective function evaluations per candidate solution.
All three methods have identical evaluation budgets and population sizes.

We perform $20$ independent runs. Each method has a population size of $20$ and proceeds for a total of $200$ generations, or $200n$ for \texttt{NSGA-II single}. Our $q$-dominance method and the NSGA-II-mean method use $n=64$ samples per candidate, where we let $n_R=n_S=8$ for the center-outward empirical quantile maps. 
Lastly, for the noise, we take $\sigma=0.1$. This choice was based on a visual inspection of the stochastic objective functions, with varied $\sigma$, evaluated at the deterministic optima. Further details, along with the plots used to determine $\sigma$, can be found in Appendix~\ref{apx:epx-zdt}.

In Figure~\ref{fig:zdt-averageHV} we show the difference in HV between the deterministic Pareto front and the expected HV, including the $95\%$ confidence interval, of the approximated fronts obtained by the different methods. Plots showing the $q\approx 0.55$ center-outward quantile regions of the final Pareto sets are shown in Appendix~\ref{apx:epx-zdt}.

We observe that for all problems, except ZDT4, the $q$-dominance method converges significantly faster than the two baselines. For ZDT4, we note that none of the solutions are close to the deterministic front in terms of the second objective, $f_2$, as can be seen in Figure~\ref{fig:apx-zdt-paretocontours} in Appendix~\ref{apx:epx-zdt}. Based on the initial inspection of the stochastic objective functions, we suspect that this is due to the sensitivity of ZDT4 to noise.

\section{Conclusion}\label{sec:conclusion}
In this paper, we presented \emph{center-outward $q$-dominance}, a sample-computable proxy for strong FSD, applicable to stochastic multi-objective optimization problems.  Our main theoretical result establishes that $q$-dominance over the full range of quantiles is \emph{sufficient} for strong FSD:
\begin{equation*}
    P_1 \succeq_q P_2,\ \forall q\in[0,1) \implies P_1 \succeq_1 P_2\ \text{(strong FSD)}.
\end{equation*}
Because each distribution is mapped once to the common uniform reference on the unit ball, the method provides globally coherent comparisons, scales to many distributions with only one optimal transport map needing to be computed per distribution, and is free of tunable hyperparameters.
The same computation yields the entire hierarchy of relaxations, for different quantiles $q$, at no extra cost, enabling a fast sorting procedure for stochastic multi-objective optimization, and our finite-sample analysis supplies an explicit sample size threshold $n^*(\delta)$ for any desired Type I error.

Our two empirical studies illustrated these properties by demonstrating the potential to enable stable rankings when traditional methods fail, and by providing faster convergence when directly applied to optimization problems.
We emphasize that these results are intended to illustrate and validate our theoretical findings, rather than to establish new state-of-the-art performance. Accordingly, we compare only with a small set of widely used stochastic baselines. A thorough benchmarking comparison of $q$-dominance against a broader range of modern techniques is beyond the scope of this paper and is left for future work.

Furthermore, we plan to compare our construction theoretically to a recent work~\cite{Rioux2025}, which is based on the (non-quadratic) optimal transport (OT) between two multivariate distributions, while our method uses the quadratic OT from each distribution to a common reference on the unit ball. In our method, the composition of the OTs $\mathbf{Q}^\pm_1\circ \mathbf{F}^\pm_2$ is unnecessarily an OT between two distributions. Although both works imply the first-order stochastic dominance, it is unclear whether our $q$-dominance implies their construction, vice versa, or if they are incompatible.

\section*{Acknowledgements}
This research is supported by the HyTROS project, funded out of the Dutch Growth Fund Program GroenvermogenNL (Green capacity for the Dutch economy and society) via the NWO call “NGF: Transport en opslag van waterstof Groenvermogen NL - Werkpakket 2”. This funding is further complemented with in-kind and cash funding from several of the HyTROS participants. 

\bibliography{references}

@book{Sriboonchita2009,
  title = {Stochastic Dominance and Applications to Finance,  Risk and Economics},
  ISBN = {9781420082678},
  url = {http://dx.doi.org/10.1201/9781420082678},
  DOI = {10.1201/9781420082678},
  publisher = {Chapman and Hall/CRC},
  author = {Sriboonchita,  Songsak and Wong,  Wing-Keung and Dhompongsa,  Sompong and Nguyen,  Hung T.},
  year = {2009},
  month = oct 
}

@article{Kopa2018StrongAW,
  title={Strong and Weak Multivariate First-Order Stochastic Dominance},
  author={Milo{\v{s}} Kopa and Barbora Petrov{\'a}},
  journal={Mathematics eJournal},
  year={2018},
  url={https://api.semanticscholar.org/CorpusID:233757705}
}

@article{McCann1995,
author = {Robert J. McCann},
title = {{Existence and uniqueness of monotone measure-preserving maps}},
volume = {80},
journal = {Duke Mathematical Journal},
number = {2},
publisher = {Duke University Press},
pages = {309 -- 323},
year = {1995},
doi = {10.1215/S0012-7094-95-08013-2},
URL = {https://doi.org/10.1215/S0012-7094-95-08013-2}
}

@TechReport{Hallin2017,
  author={Marc Hallin},
  title={{On Distribution and Quantile Functions, Ranks and Signs in R\_d}},
  year=2017,
  month=Sep,
  institution={ULB -- Universite Libre de Bruxelles},
  type={Working Papers ECARES},
  url={https://ideas.repec.org/p/eca/wpaper/2013-258262.html},
  number={ECARES 2017-34},
  doi={},
}

@article{Hallin2021,
author = {Marc Hallin and Eustasio del Barrio and Juan Cuesta-Albertos and Carlos Matr{\'a}n},
title = {{Distribution and quantile functions, ranks and signs in dimension d: A measure transportation approach}},
volume = {49},
journal = {The Annals of Statistics},
number = {2},
publisher = {Institute of Mathematical Statistics},
pages = {1139 -- 1165},
year = {2021},
doi = {10.1214/20-AOS1996},
URL = {https://doi.org/10.1214/20-AOS1996}
}

@article{ZDT,
    author = {Zitzler, Eckart and Deb, Kalyanmoy and Thiele, Lothar},
    title = {Comparison of Multiobjective Evolutionary Algorithms: Empirical Results},
    journal = {Evolutionary Computation},
    volume = {8},
    number = {2},
    pages = {173-195},
    year = {2000},
    month = {06},
    issn = {1063-6560},
    doi = {10.1162/106365600568202},
    url = {https://doi.org/10.1162/106365600568202},
    eprint = {https://direct.mit.edu/evco/article-pdf/8/2/173/1493199/106365600568202.pdf},
}

@ARTICLE{nsga2,
  author={Deb, K. and Pratap, A. and Agarwal, S. and Meyarivan, T.},
  journal={IEEE Transactions on Evolutionary Computation}, 
  title={A fast and elitist multiobjective genetic algorithm: NSGA-II}, 
  year={2002},
  volume={6},
  number={2},
  pages={182-197},
  doi={10.1109/4235.996017}}

@InProceedings{10.1007/3-540-44719-9_22,
author="Teich, J{\"u}rgen",
editor="Zitzler, Eckart
and Thiele, Lothar
and Deb, Kalyanmoy
and Coello Coello, Carlos Artemio
and Corne, David",
title="Pareto-Front Exploration with Uncertain Objectives",
booktitle="Evolutionary Multi-Criterion Optimization",
year="2001",
publisher="Springer Berlin Heidelberg",
address="Berlin, Heidelberg",
pages="314--328",
isbn="978-3-540-44719-1"
}

@inproceedings{Rioux2025,
author = {Rioux, Gabriel and Nitsure, Apoorva and Rigotti, Mattia and Greenewald, Kristjan and Mroueh, Youssef},
title = {Multivariate stochastic dominance via optimal transport and applications to models benchmarking},
year = {2025},
isbn = {9798331314385},
publisher = {Curran Associates Inc.},
address = {Red Hook, NY, USA},
booktitle = {Proceedings of the 38th International Conference on Neural Information Processing Systems},
articleno = {1237},
numpages = {34},
location = {Vancouver, BC, Canada},
series = {NIPS '24}
}

@article{Abdelaziz2012,
author = {Abdelaziz, Fouad},
year = {2012},
month = {01},
pages = {1-16},
title = {Solution approaches for the multiobjective stochastic programming},
volume = {216},
journal = {European Journal of Operational Research},
doi = {10.1016/j.ejor.2011.03.033}
}

@article{CABALLERO2004633,
title = {Stochastic approach versus multiobjective approach for obtaining efficient solutions in stochastic multiobjective programming problems},
journal = {European Journal of Operational Research},
volume = {158},
number = {3},
pages = {633-648},
year = {2004},
issn = {0377-2217},
doi = {https://doi.org/10.1016/S0377-2217(03)00371-0},
url = {https://www.sciencedirect.com/science/article/pii/S0377221703003710},
author = {Rafael Caballero and Emilio Cerdá and Marı́a {del Mar Muñoz} and Lourdes Rey}
}

@Article{Fliege2011,
author={Fliege, J{\"o}rg
and Xu, Huifu},
title={Stochastic Multiobjective Optimization: Sample Average Approximation and Applications},
journal={Journal of Optimization Theory and Applications},
year={2011},
month={Oct},
day={01},
volume={151},
number={1},
pages={135-162},
issn={1573-2878},
doi={10.1007/s10957-011-9859-6},
url={https://doi.org/10.1007/s10957-011-9859-6}
}

@article{EHRGOTT201417,
title = {Minmax robustness for multi-objective optimization problems},
journal = {European Journal of Operational Research},
volume = {239},
number = {1},
pages = {17-31},
year = {2014},
issn = {0377-2217},
doi = {https://doi.org/10.1016/j.ejor.2014.03.013},
url = {https://www.sciencedirect.com/science/article/pii/S0377221714002276},
author = {Matthias Ehrgott and Jonas Ide and Anita Schöbel}
}

@ARTICLE{1438403,
  author={Yaochu Jin and Branke, J.},
  journal={IEEE Transactions on Evolutionary Computation}, 
  title={Evolutionary optimization in uncertain environments-a survey}, 
  year={2005},
  volume={9},
  number={3},
  pages={303-317},
  doi={10.1109/TEVC.2005.846356}
}

@inproceedings{DaultonCBOZB22,
  author       = {Samuel Daulton and
                  Sait Cakmak and
                  Maximilian Balandat and
                  Michael A. Osborne and
                  Enlu Zhou and
                  Eytan Bakshy},
  editor       = {Kamalika Chaudhuri and
                  Stefanie Jegelka and
                  Le Song and
                  Csaba Szepesv{\'{a}}ri and
                  Gang Niu and
                  Sivan Sabato},
  title        = {{Robust Multi-Objective Bayesian Optimization Under Input Noise}},
  booktitle    = {International Conference on Machine Learning, {ICML} 2022, 17-23 July
                  2022, Baltimore, Maryland, {USA}},
  series       = {Proceedings of Machine Learning Research},
  volume       = {162},
  pages        = {4831--4866},
  publisher    = {{PMLR}},
  year         = {2022},
  url          = {https://proceedings.mlr.press/v162/daulton22a.html},
  bibsource    = {dblp computer science bibliography, https://dblp.org}
}

@article{trappler2025multiobjective,
  title={{Multiobjective Optimization under Uncertainties using Conditional Pareto Fronts}},
  author={Trappler, Victor and Helbert, C{\'e}line and Riche, Rodolphe Le},
  journal={arXiv preprint arXiv:2504.04944},
  year={2025}
}

@misc{manole2024pluginestimationsmoothoptimal,
      title={Plugin Estimation of Smooth Optimal Transport Maps}, 
      author={Tudor Manole and Sivaraman Balakrishnan and Jonathan Niles-Weed and Larry Wasserman},
      year={2024},
      eprint={2107.12364},
      archivePrefix={arXiv},
      primaryClass={math.ST},
      url={https://arxiv.org/abs/2107.12364}, 
}

@InProceedings{pfisterer22a,
  title = 	 {YAHPO Gym - An Efficient Multi-Objective Multi-Fidelity Benchmark for Hyperparameter Optimization},
  author =       {Pfisterer, Florian and Schneider, Lennart and Moosbauer, Julia and Binder, Martin and Bischl, Bernd},
  booktitle = 	 {Proceedings of the First International Conference on Automated Machine Learning},
  pages = 	 {3/1--39},
  year = 	 {2022},
  editor = 	 {Guyon, Isabelle and Lindauer, Marius and van der Schaar, Mihaela and Hutter, Frank and Garnett, Roman},
  volume = 	 {188},
  series = 	 {Proceedings of Machine Learning Research},
  month = 	 {25--27 Jul},
  publisher =    {PMLR},
  url = 	 {https://proceedings.mlr.press/v188/pfisterer22a.html}
}

@ARTICLE{ParEGO,
  author={Knowles, J.},
  journal={IEEE Transactions on Evolutionary Computation}, 
  title={ParEGO: a hybrid algorithm with on-line landscape approximation for expensive multiobjective optimization problems}, 
  year={2006},
  volume={10},
  number={1},
  pages={50-66},
  doi={10.1109/TEVC.2005.851274}}

@InProceedings{SMS-EGO,
author="Ponweiser, Wolfgang
and Wagner, Tobias
and Biermann, Dirk
and Vincze, Markus",
editor="Rudolph, G{\"u}nter
and Jansen, Thomas
and Beume, Nicola
and Lucas, Simon
and Poloni, Carlo",
title="Multiobjective Optimization on a Limited Budget of Evaluations Using Model-Assisted {$\mathcal{S}$}-Metric Selection",
booktitle="Parallel Problem Solving from Nature -- PPSN X",
year="2008",
publisher="Springer Berlin Heidelberg",
address="Berlin, Heidelberg",
pages="784--794",
isbn="978-3-540-87700-4"
}

@ARTICLE{EHVI,
  author={Emmerich, M.T.M. and Giannakoglou, K.C. and Naujoks, B.},
  journal={IEEE Transactions on Evolutionary Computation}, 
  title={Single- and multiobjective evolutionary optimization assisted by Gaussian random field metamodels}, 
  year={2006},
  volume={10},
  number={4},
  pages={421-439},
  doi={10.1109/TEVC.2005.859463}}

@INPROCEEDINGS{MEGO,
  author={Jeong, S. and Obayashi, S.},
  booktitle={2005 IEEE Congress on Evolutionary Computation}, 
  title={Efficient global optimization (EGO) for multi-objective problem and data mining}, 
  year={2005},
  volume={3},
  number={},
  pages={2138-2145 Vol. 3},
  doi={10.1109/CEC.2005.1554959}}

@ARTICLE{MIES,
  author={Li, Rui and Emmerich, Michael T.M. and Eggermont, Jeroen and Bäck, Thomas and Schütz, M. and Dijkstra, J. and Reiber, J.H.C.},
  journal={Evolutionary Computation}, 
  title={Mixed Integer Evolution Strategies for Parameter Optimization}, 
  year={2013},
  volume={21},
  number={1},
  pages={29-64},
  doi={10.1162/EVCO_a_00059}
}

@inproceedings{DruganN13,
  author       = {Madalina M. Drugan and
                  Ann Now{\'{e}}},
  title        = {Designing multi-objective multi-armed bandits algorithms: {A} study},
  booktitle    = {The 2013 International Joint Conference on Neural Networks, {IJCNN}
                  2013, Dallas, TX, USA, August 4-9, 2013},
  pages        = {1--8},
  publisher    = {{IEEE}},
  year         = {2013},
  url          = {https://doi.org/10.1109/IJCNN.2013.6707036},
  doi          = {10.1109/IJCNN.2013.6707036},
  bibsource    = {dblp computer science bibliography, https://dblp.org}
}

@inproceedings{XuK23,
  author       = {Mengfan Xu and
                  Diego Klabjan},
  editor       = {Andreas Krause and
                  Emma Brunskill and
                  Kyunghyun Cho and
                  Barbara Engelhardt and
                  Sivan Sabato and
                  Jonathan Scarlett},
  title        = {{Pareto Regret Analyses in Multi-objective Multi-armed Bandit}},
  booktitle    = {International Conference on Machine Learning, {ICML} 2023, 23-29 July
                  2023, Honolulu, Hawaii, {USA}},
  series       = {Proceedings of Machine Learning Research},
  volume       = {202},
  pages        = {38499--38517},
  publisher    = {{PMLR}},
  year         = {2023},
  url          = {https://proceedings.mlr.press/v202/xu23i.html},
  bibsource    = {dblp computer science bibliography, https://dblp.org}
}

@inproceedings{LuWHZ19,
  author       = {Shiyin Lu and
                  Guanghui Wang and
                  Yao Hu and
                  Lijun Zhang},
  editor       = {Sarit Kraus},
  title        = {{Multi-Objective Generalized Linear Bandits}},
  booktitle    = {Proceedings of the Twenty-Eighth International Joint Conference on
                  Artificial Intelligence, {IJCAI} 2019, Macao, China, August 10-16,
                  2019},
  pages        = {3080--3086},
  publisher    = {ijcai.org},
  year         = {2019},
  url          = {https://doi.org/10.24963/ijcai.2019/427},
  doi          = {10.24963/IJCAI.2019/427},
  bibsource    = {dblp computer science bibliography, https://dblp.org}
}

@Inbook{stochasticorder-book,
editor="Shaked, Moshe
and Shanthikumar, J. George",
title="Multivariate Stochastic Orders",
bookTitle="Stochastic Orders",
year="2007",
publisher="Springer New York",
address="New York, NY",
pages="265--322",
isbn="978-0-387-34675-5",
doi="10.1007/978-0-387-34675-5_6",
url="https://doi.org/10.1007/978-0-387-34675-5_6"
}

@ARTICLE{pymoo,
    author={J. {Blank} and K. {Deb}},
    journal={IEEE Access},
    title={pymoo: Multi-Objective Optimization in Python},
    year={2020},
    volume={8},
    number={},
    pages={89497-89509},
}

\appendix
\section{Proof of Theorem~\ref{thm:pacbound}} \label{apx:proof-pacthm}
Let $\mathcal{G}_n\coloneqq\left\{q\mathbf{u} : (\mathbf{u},q)\in\mathfrak{S}^{(n_S)} \times \left\{\frac{k}{n_R+1}\right\}_{k=1}^{n_R}\right\}$ be the set of grid points on the augmented grid. Here we take $n_0=0$ for the augmented grid for simplicity; for $n_0>0$, a similar argument suffices.
Furthermore, define
\begin{align*}
    \Delta &\coloneqq \inf_{\mathbf{z}\in\mathbb{S}^d}\min_{k=1,\dots,d} \left( \mathbf{Q}^\pm_1(\mathbf{z}) - \mathbf{Q}^\pm_2(\mathbf{z}) \right)_k,\quad \mathrm{and}\\
    \widehat{\Delta} &\coloneqq \min_{\mathbf{z}\in\mathcal{G}_n}\min_{k=1,\dots,d} \left( \widehat{\mathbf{Q}}^\pm_1(\mathbf{z}) - \widehat{\mathbf{Q}}^\pm_2(\mathbf{z}) \right)_k,
\end{align*}
then $P_1 \succeq_q P_2$ for every $q\in [0,1)$ if and only if $\Delta \geq 0$, and $\mathbf{X}^{(n)} \succeq_q \mathbf{Y}^{(n)}$ for every $q\in\left\{\frac{k}{n_R+1}\right\}_{k=1}^{n_R}$ if and only if $\widehat{\Delta} \geq 0$.

Now if we have that $\Delta \geq 0$ and
\begin{equation*}
    \max_{i=1,2} \max_{\mathbf{z}\in\mathcal{G}_n} \left\lVert \widehat{\mathbf{Q}}^\pm_i(\mathbf{z}) - \mathbf{Q}^\pm_i(\mathbf{z}) \right\rVert_\infty \leq \frac{\Delta}{2},
\end{equation*}
then we must also have that $\widehat{\Delta}\geq 0$.
Since, 
\begin{align*}
    &\left| \left( \widehat{\mathbf{Q}}^\pm_1(\mathbf{z}) - \widehat{\mathbf{Q}}^\pm_2(\mathbf{z}) \right)_k - \left( \mathbf{Q}^\pm_1(\mathbf{z}) - \mathbf{Q}^\pm_2(\mathbf{z}) \right)_k \right|\\ 
    &= \left| \left( \widehat{\mathbf{Q}}^\pm_1(\mathbf{z}) - \mathbf{Q}^\pm_1(\mathbf{z}) \right)_k - \left( \widehat{\mathbf{Q}}^\pm_2(\mathbf{z}) - \mathbf{Q}^\pm_2(\mathbf{z}) \right)_k \right|\\
    &\leq \Delta,
\end{align*}
which implies that
\begin{equation*}
    \left( \widehat{\mathbf{Q}}^\pm_1(\mathbf{z}) - \widehat{\mathbf{Q}}^\pm_2(\mathbf{z}) \right)_k \geq \left( \mathbf{Q}^\pm_1(\mathbf{z}) - \mathbf{Q}^\pm_2(\mathbf{z}) \right)_k - \Delta \geq 0.
\end{equation*}
Thus, we only need to prove that 
\begin{equation*}
    \max_{i=1,2} \max_{\mathbf{z}\in\mathcal{G}_n} \left\lVert \widehat{\mathbf{Q}}^\pm_i(\mathbf{z}) - \mathbf{Q}^\pm_i(\mathbf{z}) \right\rVert_\infty \leq \frac{\Delta}{2}.
\end{equation*}

By Corollary 3.2 of \cite{Hallin2021}, there exists a smooth interpolation $\overline{\mathbf{Q}}^\pm_i$ of $\widehat{\mathbf{Q}}^\pm_i$ that is $\overline{L}_i$-Lipschitz continuous, and agrees with the empirical map at each point on the augmented grid, that is $\overline{\mathbf{Q}}^\pm_i(\mathbf{z})=\widehat{\mathbf{Q}}^\pm_i(\mathbf{z})$ for all $\mathbf{z}\in\mathcal{G}_n$.
Furthermore, by assumption $\mathbf{Q}^\pm_i$ is $L_i$-bi-Lipschitz, and thus
\begin{align*}
    \left\lVert \overline{\mathbf{Q}}^\pm_i(\mathbf{x}) - \overline{\mathbf{Q}}^\pm_i(\mathbf{y}) \right\rVert_2 &\leq \overline{L}_i\lVert \mathbf{x}-\mathbf{y}\rVert_2, \\
    \left\lVert \mathbf{Q}^\pm_i(\mathbf{x}) - \mathbf{Q}^\pm_i(\mathbf{y}) \right\rVert_2 &\leq L_i \lVert \mathbf{x}-\mathbf{y}\rVert_2.
\end{align*}
Using these properties and inequalities, together with the triangle inequality and the fact that $\lVert \mathbf{x}\rVert_\infty \leq \lVert \mathbf{x}\rVert_2$ for all $\mathbf{x}$, we get that
\begin{align*}
    \left\lVert \widehat{\mathbf{Q}}^\pm_i(\mathbf{z}) - \mathbf{Q}^\pm_i(\mathbf{z}) \right\rVert_\infty &= \left\lVert \overline{\mathbf{Q}}^\pm_i(\mathbf{z}) - \mathbf{Q}^\pm_i(\mathbf{z}) \right\rVert_\infty\\
    &\leq \left\lVert \overline{\mathbf{Q}}^\pm_i(\mathbf{z}) - \mathbf{Q}^\pm_i(\mathbf{z}) \right\rVert_2 \\
    &\leq \left\lVert \overline{\mathbf{Q}}^\pm_i(\mathbf{v}) - \mathbf{Q}^\pm_i(\mathbf{v}) \right\rVert_2 \\
    &\phantom{\leq\lVert}+ L_i \lVert\mathbf{z}-\mathbf{v}\rVert_2 + \overline{L}_i \lVert\mathbf{z}-\mathbf{v}\rVert_2,
\end{align*}
for any $\mathbf{z}\in\mathcal{G}$ and $\mathbf{v}\in\mathbb{S}_d$.

Now select any maximal $\epsilon_\mathrm{net}$-separated set $\mathcal{N}_{\epsilon_\mathrm{net}}\subset\mathbb{S}^d$. Such a set is automatically an $\epsilon_\mathrm{net}$-net, that is for any $\mathbf{z}\in\mathbb{S}^d$ there exists a $\mathbf{v}\in\mathcal{N}_{\epsilon_\mathrm{net}}$ such that $\lVert \mathbf{z} - \mathbf{v}\rVert_2 \leq \epsilon_\mathrm{net}$. By a standard volume argument we have that $|\mathcal{N}_{\epsilon_\mathrm{net}}|\leq c_d\epsilon_\mathrm{net}^{-d}$.
If we choose $\epsilon_\mathrm{net} \coloneqq \frac{\Delta}{4(L_i+\overline{L}_i)}$, we get that for all $\mathbf{z}\in\mathbb{S}^d$ there exists a $\mathbf{v}\in\mathcal{N}_{\epsilon_\mathrm{net}}$ such that

\begin{align*}
    \left\lVert \widehat{\mathbf{Q}}^\pm_i(\mathbf{z}) - \mathbf{Q}^\pm_i(\mathbf{z}) \right\rVert_\infty &\leq \left\lVert \overline{\mathbf{Q}}^\pm_i(\mathbf{v}) - \mathbf{Q}^\pm_i(\mathbf{v}) \right\rVert_2\\
    &\phantom{\leq\lVert} + (L_i + \overline{L}_i) \epsilon_\mathrm{net}\\
    &\leq \left\lVert \overline{\mathbf{Q}}^\pm_i(\mathbf{v}) - \mathbf{Q}^\pm_i(\mathbf{v}) \right\rVert_2 + \frac{\Delta}{4}.
\end{align*}
It follows that
\begin{gather*}
    \mathbb{P}\left(\max_{\mathbf{z}\in\mathcal{G}_n} \left\lVert \widehat{\mathbf{Q}}^\pm_i(\mathbf{z}) - \mathbf{Q}^\pm_i(\mathbf{z}) \right\rVert_\infty \leq \frac{\Delta}{2}\right)\\ \geq\\
    \mathbb{P}\left(\max_{\mathbf{v}\in\mathcal{N}_{\epsilon_\mathrm{net}}} \left\lVert \overline{\mathbf{Q}}^\pm_i(\mathbf{v}) - \mathbf{Q}^\pm_i(\mathbf{v}) \right\rVert_2 \leq \frac{\Delta}{4}\right).
\end{gather*}

Chebyshev's inequality yields
\begin{gather*}
    \rho\left(\left\{ \mathbf{z} : \left\lVert \overline{\mathbf{Q}}^\pm_i(\mathbf{z}) - \mathbf{Q}^\pm_i(\mathbf{z}) \right\rVert_2 > \frac{\Delta}{4} \right\}\right) 
    \\ \leq \\
    \frac{16}{\Delta^2} \left\lVert \overline{\mathbf{Q}}_i^\pm - \mathbf{Q}_i^\pm \right\rVert^2_{L^2(U_d)},
\end{gather*}
where $U_d$ is the uniform measure over $\mathbb{S}^d$, and thus by using a union bound we get
\begin{gather*}
    \mathbb{P}\left(\max_{\mathbf{v}\in\mathcal{N}_{\epsilon_\mathrm{net}}} \left\lVert \overline{\mathbf{Q}}^\pm_i(\mathbf{v}) - \mathbf{Q}^\pm_i(\mathbf{v}) \right\rVert_2 \leq \frac{\Delta}{4} \right) 
    \\ \geq \\
    1 - \frac{16 c_d\epsilon_\mathrm{net}^{-d}}{\Delta^2} \left\lVert \overline{\mathbf{Q}}_i^\pm - \mathbf{Q}_i^\pm \right\rVert^2_{L^2(U_d)}.
\end{gather*}

All that remains is to upper bound $\lVert\overline{\mathbf{Q}}_i^\pm - \mathbf{Q}_i^\pm \rVert^2_{L^2(U_d)}$. Unfortunately, $\overline{\mathbf{Q}}_i^\pm$ is not an optimal transport map, and therefore, upper bounds from the literature do not apply here. Instead we first upper bound the term by $\lVert\widehat{\mathbf{Q}}_i^\pm - \mathbf{Q}_i^\pm \rVert^2_{L^2\left(\widehat{U}_d\right)}$, where $\widehat{U}_d=\frac{1}{n}\sum_{\mathbf{z}\in\mathcal{G}_n} \delta_\mathbf{z}$ which weakly converges to $U_d$. 

For this we consider a Voronoi partition of $\mathbb{S}^d$, that is each grid point $\mathbf{z}\in\mathcal{G}_n$ has a corresponding cell 
\begin{equation*}
    V(\mathbf{z}) = \left\{ \mathbf{x}\in\mathbb{S}^d : \lVert \mathbf{z}-\mathbf{x}\rVert_2 \leq \lVert \mathbf{z}'-\mathbf{x}\rVert_2, \forall \mathbf{z}'\in\mathcal{G}_n\right\}.
\end{equation*}
By the construction of the grid, we know that each cell has a radius of at most $h_n = C_d \max(n_R^{-1}, n_S^{-1/(d-1)})$, and thus for every $\mathbf{x}\in V(\mathbf{z})$ we have that
\begin{equation*}
    \left\lVert \overline{\mathbf{Q}}_i^\pm(\mathbf{x}) - \mathbf{Q}_i^\pm(\mathbf{x}) \right\rVert_2^2 \leq \left\lVert \widehat{\mathbf{Q}}_i^\pm(\mathbf{z}) - \mathbf{Q}_i^\pm(\mathbf{z}) \right\rVert_2^2 + \overline{L}_i^2h_n^2,
\end{equation*}
where we again used the $\overline{L}_i$-Lipschitz continuity of $\overline{\mathbf{Q}}_i^\pm$.
Using this inequality, we get
\begin{align*}
    \lVert\overline{\mathbf{Q}}&_i^\pm - \mathbf{Q}_i^\pm \rVert^2_{L^2(U_d)}\\
    &=\int_{\mathbb{S}^d} \left\lVert \overline{\mathbf{Q}}_i^\pm(\mathbf{x}) - \mathbf{Q}_i^\pm(\mathbf{x}) \right\rVert_2^2\ \mathrm{d}U_d(x) \\
    &\leq \int_{\mathbb{S}^d} \left\lVert \widehat{\mathbf{Q}}_i^\pm(\sigma(\mathbf{x})) - \mathbf{Q}_i^\pm(\sigma(\mathbf{x})) \right\rVert_2^2 + \overline{L}_i^2h_n^2\ \mathrm{d}U_d(x) \\
    &= \int_{\mathbb{S}^d} \left\lVert \widehat{\mathbf{Q}}_i^\pm(\sigma(\mathbf{x})) - \mathbf{Q}_i^\pm(\sigma(\mathbf{x})) \right\rVert_2^2\ \mathrm{d}U_d(x) + \overline{L}_i^2h_n^2 \\
    &= \overline{L}_i^2h_n^2 + \sum_{\mathbf{z}\in\mathcal{G}_n} U_d\left(V_\mathbf{z}\right) \left\lVert \widehat{\mathbf{Q}}_i^\pm(\mathbf{z}) - \mathbf{Q}_i^\pm(\mathbf{z}) \right\rVert_2^2,
\end{align*}
where $\sigma: \mathbb{S}^d\to\mathcal{G}$ is the function that maps each $\mathbf{x}\in V(\mathbf{z})$ to the grid point $\mathbf{z}\in\mathcal{G}_n$. We can conservatively upper bound $U_d(V_\mathbf{z}) \leq h_n^d$, as this is the ratio between a ball with radius $h_n$ and the unit ball.
\begin{align*}
    \lVert\overline{\mathbf{Q}}&_i^\pm - \mathbf{Q}_i^\pm \rVert^2_{L^2(U_d)}\\
    &\leq \overline{L}_i^2h_n^2 + h_n^d\sum_{\mathbf{z}\in\mathcal{G}_n} \left\lVert \widehat{\mathbf{Q}}_i^\pm(\mathbf{z}) - \mathbf{Q}_i^\pm(\mathbf{z}) \right\rVert_2^2\\
    &= \overline{L}_i^2 h^2_n + nh_n^d \left\lVert\widehat{\mathbf{Q}}_i^\pm - \mathbf{Q}_i^\pm \right\rVert^2_{L^2\left(\widehat{U}_d\right)}
\end{align*}
By Proposition 14 of \cite{manole2024pluginestimationsmoothoptimal} we have that
\begin{equation*}
    \mathbb{E}\left\lVert\widehat{\mathbf{Q}}_i^\pm - \mathbf{Q}_i^\pm \right\rVert^2_{L^2\left(\widehat{U}_d\right)} \lesssim \kappa_n \coloneqq \begin{cases}
        n^{-1/2}, & d\leq 3\\ n^{-1/2}\log n, & d=4\\ n^{-2/d}, & d \geq 5
    \end{cases},
\end{equation*}
Let $n_R=n^\theta$ and $n_S=n^{1-\theta}$, then 
\begin{equation*}
    h_n = C_d \begin{cases}
        n^{-\theta}, & \theta \leq \frac{1}{d}\\
        n^{-\frac{1-\theta}{d-1}}, & \theta > \frac{1}{d}
    \end{cases}.
\end{equation*}
We require that $1-\theta +\beta(d) <0$ for $\theta \leq \frac{1}{d}$ and $1 - \frac{1-\theta}{d-1} +\beta(d) <0$ for $\theta > \frac{1}{d}$, where $\beta(d)$ is the exponent of $n$ in the $\kappa_n$ term ignoring any $\log$ factors.
This yields the following viable intervals for $\theta$:
\begin{equation*}
    \theta \in \begin{cases}
        \left(\frac{1}{2d}, \frac{d+1}{2d}\right), & d\leq 4\\
        \left(\frac{d-2}{d^2}, \frac{2d-3}{d^2}\right), & d\geq 5
    \end{cases}.
\end{equation*}

We now need to include the expectation bound in the final inequality. To do this consider the events $F_i\coloneqq \left\{ \max_{\mathbf{v}\in\mathcal{N}_{\epsilon_\mathrm{net}}} \left\lVert \overline{\mathbf{Q}}^\pm_i(\mathbf{v}) - \mathbf{Q}^\pm_i(\mathbf{v}) \right\rVert_2 \leq \frac{\Delta}{4} \right\}$ and $A\coloneq \left\{ \left\lVert\overline{\mathbf{Q}}_i^\pm - \mathbf{Q}_i^\pm \right\rVert^2_{L^2(U_d)} \leq t \right\}$, then Markov's inequality yields
\begin{equation*}
    \mathbb{P}(A^c) \leq \frac{1}{t} \mathbb{E}\left\lVert\overline{\mathbf{Q}}_i^\pm - \mathbf{Q}_i^\pm \right\rVert^2_{L^2(U_d)}.
\end{equation*}
Furthermore, we have that
\begin{equation*}
    \mathbb{P}(F_i^c\ \vert\ A) \leq \frac{16 c_d\epsilon_\mathrm{net}^{-d}}{\Delta^2}\left\lVert\overline{\mathbf{Q}}_i^\pm - \mathbf{Q}_i^\pm \right\rVert^2_{L^2({U}_d)}.
\end{equation*}
By the law of total probability we have that $\mathbb{P}(F_i^c) = \mathbb{P}(F_i^c\ \vert\ A)\mathbb{P}(A) + \mathbb{P}(F_i^c\ \vert\ A^c)\mathbb{P}(A^c)$. Using the worst-case bounds $\mathbb{P}(A), \mathbb{P}(F_i^c\ \vert\ A^c) \leq 1$ we get
\begin{align*}
    \mathbb{P}(F_i^c) &\leq \mathbb{P}(F_i^c) + \mathbb{P}(A^c)\\
    &\leq \frac{16 c_d\epsilon_\mathrm{net}^{-d}}{\Delta^2}t + \frac{1}{t} \mathbb{E}\left\lVert\overline{\mathbf{Q}}_i^\pm - \mathbf{Q}_i^\pm \right\rVert^2_{L^2(U_d)},
\end{align*}
taking $t=\sqrt{\frac{\Delta^2}{16c_d\epsilon_\mathrm{net}^{-d}}\mathbb{E}\left\lVert\overline{\mathbf{Q}}_i^\pm - \mathbf{Q}_i^\pm \right\rVert^2_{L^2(U_d)}}$ yields
\begin{equation*}
    \mathbb{P}(F_i^c) \leq \frac{8\sqrt{c_d\epsilon_\mathrm{net}^{-d}}}{\Delta}\left( \mathbb{E}\left\lVert\overline{\mathbf{Q}}_i^\pm - \mathbf{Q}_i^\pm \right\rVert^2_{L^2(U_d)} \right)^{1/2}.
\end{equation*}
Putting everything together, we get
\begin{gather*}
    \mathbb{P}\left(\max_{i=1,2}\max_{\mathbf{z}\in\mathcal{G}} \left\lVert \widehat{\mathbf{Q}}^\pm_i(\mathbf{z}) - \mathbf{Q}^\pm_i(\mathbf{z}) \right\rVert_\infty \leq \frac{\Delta}{2}\right)
    \\ \geq \\
    1 - \sum_{i=1}^2\frac{8\sqrt{c_d\epsilon_\mathrm{net}^{-d}}}{\Delta}\left( \overline{L}_i^2 h^2_n + C_i n h_n^d \kappa_n \right)^{1/2}
    \\ = \\
    1 - \sum_{i=1}^2 \frac{8\sqrt{c_d}(4[L_i+\overline{L}_i])^{d/2}}{\Delta^{1+d/2}}\left( \overline{L}_i^2 h^2_n + C_i n h_n^d \kappa_n \right)^{1/2}
\end{gather*}
We can solve
\begin{equation*}
    \delta \geq  \sum_{i=1}^2 \frac{8\sqrt{c_d}(4[L_i+\overline{L}_i])^{d/2}}{\Delta^{1+d/2}}\left( \overline{L}_i^2 h^2_n + C_i n h_n^d \kappa_n \right)^{1/2},
\end{equation*}
for $n$ to find our threshold $n^*$. However, we cannot do this directly, because the sum and the two terms in the square root both depend on $n$. 
We first get rid of the sum by letting $L=\max\{L_1, L_2\}$, $\overline{L}=\max\{\overline{L}_1, \overline{L}_2\}$, and $C=\max\{C_1, C_2\}$ and multiplying the RHS by 2, giving
\begin{equation*}
    \delta \geq \frac{16\sqrt{c_d}(4[L+\overline{L}])^{d/2}}{\Delta^{1+d/2}}\left( \overline{L}^2 h^2_n + C n h_n^d \kappa_n \right)^{1/2}.
\end{equation*}
By giving both terms equal "weight" we can break this up into two inequalities
\begin{align*}
     \overline{L}^2 h_n^2 &\leq \frac{1}{2}\frac{\delta^2 \Delta^{d+2}}{256c_d(4[L+\overline{L}])^d},\\
     C n h_n^d\kappa_n &\leq \frac{1}{2}\frac{\delta^2 \Delta^{d+2}}{256c_d(4[L+\overline{L}])^d}.
\end{align*}
Then $n^* = \max\{n_1^*, n_2^*\}$, with
\begin{align*}
    n_1^* &= \min\left\{ n :  \overline{L}^2 h_n^2 \leq \frac{1}{2}\frac{\delta^2 \Delta^{d+2}}{256c_d(4[L+\overline{L}])^d}\right\},\\
    n_2^* &= \min\left\{ n :   C n h_n^d\kappa_n \leq \frac{1}{2}\frac{\delta^2 \Delta^{d+2}}{256c_d(4[L+\overline{L}])^d}\right\}.
\end{align*}
In order to find expressions for $n_1^*$ and $n_2^*$ we have to take into account all six cases; $d\geq 3$, $d=4$, and $d\geq 5$, as well as  $\theta \leq \frac{1}{d}$ and $\theta > \frac{1}{d}$. 
A calculation yields the final expression for $n^*$ shown in \eqref{eq:apx-n*}, concluding the proof. \hfill \qedsymbol
\begin{figure*}
    \centering
    \begin{equation}\label{eq:apx-n*}
    n^* = \begin{cases}
        \max\left\{ \left( \frac{512 c_d \overline{L}^2(4[L+\overline{L}])^d}{\delta^2 \Delta^{d+2}} \right)^{\frac{1}{2\theta}}, 
                    \left( \frac{\delta^2 \Delta^{d+2}}{512 c_d C (4[L+\overline{L}])^2} \right)^{\frac{2}{1-2\theta d}} \right\}, \quad d \leq 3\ \wedge\ \theta\leq\frac{1}{d},\\
        \max\left\{ \left( \frac{512 c_d \overline{L}^2(4[L+\overline{L}])^d}{\delta^2 \Delta^{d+2}} \right)^{\frac{d-1}{1-\theta}}, 
                    \left( \frac{\delta^2 \Delta^{d+2}}{512 c_d C (4[L+\overline{L}])^2} \right)^{\frac{2(d-1)}{2\theta d -d -1}} \right\}, \quad d \leq 3\ \wedge\ \theta > \frac{1}{d},\\
        \max\left\{ \left( \frac{512 c_d \overline{L}^2(4[L+\overline{L}])^d}{\delta^2 \Delta^{d+2}} \right)^{\frac{1}{2\theta}},
                    \exp\left( \frac{1}{\frac{1}{2}-\theta d} W_{-1}\left( \left[ \frac{1}{2} -\theta d \right] \frac{\delta^2 \Delta^{d+2}}{512 c_d C (4[L+\overline{L}])^2} \right) \right) \right\}, \quad d = 4\ \wedge\ \theta\leq\frac{1}{d},\\
        \max\left\{ \left( \frac{512 c_d \overline{L}^2(4[L+\overline{L}])^d}{\delta^2 \Delta^{d+2}} \right)^{\frac{d-1}{1-\theta}}, 
                    \exp\left( \frac{1}{\frac{1}{2}- d\frac{1-\theta}{d-1}} W_{-1}\left( \left[ \frac{1}{2}- d\frac{1-\theta}{d-1} \right] \frac{\delta^2 \Delta^{d+2}}{512 c_d C (4[L+\overline{L}])^2} \right) \right) \right\}, \quad d = 4\ \wedge\ \theta > \frac{1}{d},\\
        \max\left\{ \left( \frac{512 c_d \overline{L}^2(4[L+\overline{L}])^d}{\delta^2 \Delta^{d+2}} \right)^{\frac{1}{2\theta}},
                    \left( \frac{\delta^2 \Delta^{d+2}}{512 c_d C (4[L+\overline{L}])^2} \right)^{\frac{d}{d-2-\theta d^2}} \right\}, \quad d\geq 5\ \wedge\ \theta\leq\frac{1}{d},\\
        \max\left\{ \left( \frac{512 c_d \overline{L}^2(4[L+\overline{L}])^d}{\delta^2 \Delta^{d+2}} \right)^{\frac{d-1}{1-\theta}},
                    \left( \frac{\delta^2 \Delta^{d+2}}{512 c_d C (4[L+\overline{L}])^2} \right)^{\frac{d(d-1)}{\theta d^2 -3d +2}} \right\}, \quad d\geq 5\ \wedge\ \theta > \frac{1}{d}.
    \end{cases}
\end{equation}
\end{figure*}

\section{$q$-Dominance Sorting Algorithm}\label{apx:alg}
Algorithm~\ref{alg:q-sort} contains the sorting procedure using $q$-dominance. In lines 1 and 2, we compute a dominance matrix $\mathcal{D}\in[0,1]^{n_R \times m \times m}$. We iterate over lines 6 through 13 until all candidates have been assigned to a front $\mathcal{F}_l$. The assignment happens as follows: of the remaining candidates $\mathcal{C} \setminus \bigcup_{l=1}^n \mathcal{F}_l$, we assign the ones that are not dominated by any other candidate up to quantile $\frac{n_R-c}{n_R+1}$ to the front $\mathcal{F}_{n+1}$ and sort them, inside the front, based on $\sum_{k=1}^{n_R-c}\sum_{i\in\mathcal{C}} \mathcal{D}_{kij}$, which is a measure of how close candidate $j\in\mathcal{F}_{n+1}$ is to being dominated by the candidates in $\mathcal{C}$. The variable $c$ loops over all empirical quantiles, increasing by $1$ after each iteration and resetting to $0$ after it has reached $c=n_R$. When $c$ is reset and we have looped over all quantiles, we reduce the list of candidates by removing those already assigned to a front.
\begin{algorithm*}
    \caption{Sorting algorithm using $q$-dominance}\label{alg:q-sort}
    \begin{algorithmic}[1]
        \REQUIRE CO-Quantile maps $\mathbf{Q}^\pm_1, \dots, \mathbf{Q}^\pm_m$ defined on the same augmented grid.
        
        \STATE $ \mathcal{D}_{kij} \gets\frac{1}{n_S}\sum_{\mathbf{u}\in\mathfrak{S}^{(n_S)}} \mathbf{1}_{\left\{ \widehat{\mathbf{Q}}^\pm_i\left( \frac{k}{n_R+1}\mathbf{u} \right) \geq \widehat{\mathbf{Q}}^\pm_j\left( \frac{k}{n_R+1}\mathbf{u} \right) \right\}}$
        \STATE $\mathcal{D}_{kii} \gets0$
        \STATE $\mathcal{C}\gets[m]$ \qquad \COMMENT{unsorted candidates}
        \STATE $n\gets 0,\quad c\gets 0$
        \WHILE{$[m] \setminus \bigcup_{l=1}^n\mathcal{F}_l\not=\emptyset$}
        \STATE $\mathcal{F}_{n+1} \gets \left\{ j\in \mathcal{C} \setminus \bigcup_{l=1}^n \mathcal{F}_l : \mathcal{D}_{kij} \not=1,\  \forall k\in[n_R-c]\ \forall i\in \mathcal{C}  \right\}$
        \STATE Sort $j\in\mathcal{F}_{n+1}$ by increasing $\sum_{k=1}^{n_R-c}\sum_{i\in\mathcal{C}} \mathcal{D}_{kij}$.
        \IF{$c=n_R$}
            \STATE $c\gets0$
            \STATE $\mathcal{C} \gets \mathcal{C} \setminus \bigcup_{l=1}^n\mathcal{F}_l$
        \ELSE
            \STATE $c \gets c+1$
        \ENDIF
        
        \STATE $n \gets n+1$ 
        \ENDWHILE
        \ENSURE $\mathcal{F}_1,\dots,\mathcal{F}_n$
    \end{algorithmic}
\end{algorithm*}

\section{Experimental details}
All experiments were conducted using Python 3.10.8 on a laptop running Ubuntu 22.04 LTS, equipped with an Intel Core i7-12700M processor and 15 GB of RAM.
\subsection{YAHPO-MO Rankings}\label{apx:exp-hpo}
All the optimizers were run for a total budget of $\left\lceil 20 +40\sqrt{\text{SEARCH\_SPACE\_DIM}} \right\rceil$ evaluations across 30 replications. More details on the experimental setup can be found in \cite{pfisterer22a}.

Out of the 25 problem instances, which are shown in Table~\ref{tab:apx-hpo-problems}, there are six instances, three of rbv2\_super and three of rbv2\_xgboost, that appear to have some global optimum. This optimum is several orders of magnitude better than other solutions, and in five out of six cases, it is only found by one optimizer: MIES.
Since we are interested in comparing stochastic solutions, and these optima consist of deterministic points, we exclude these six instances from our analysis.

For the $q$-dominance ranking, we sample $k \leq 5$ points uniformly at random from the Pareto front at each replication, resulting in $30k$ samples, for each problem instance at each fraction of the budget used. A single random seed is set at the start of the experiment, and the random number generator's state is carried forward across successive replications.
We take $n_R \approx (30k)^{1/d}$, such that with $n_S = \frac{30k}{n_R}$, we have $30k=n_R n_S$.
The HPO methods are then ranked based on the procedure in Algorithm~\ref{alg:q-sort}, using these $30k$ samples, allowing for ties when the values computed in line 7 of Algorithm~\ref{alg:q-sort} are equal.

Along with the critical difference diagrams in Figure~\ref{fig:hpo-cd100}, we performed the corresponding Friedman tests, which indicated significant differences $p<0.001$ for both the HVI ranks and the $q$-dominance ranks.

\begin{table*}
    \centering
    \begin{tabular}{@{}llll@{}}
    \toprule
    \textbf{Scenario} & \textbf{Instance} & \textbf{Objectives}                & \textbf{Included} \\ \midrule
    iaml\_glmnet      & 1489              & mmce, nf                           & Yes               \\
    iaml\_glmnet      & 1067              & mmce, nf                           & Yes               \\
    iaml\_ranger      & 1489              & mmce, nf, ias                      & Yes               \\
    iaml\_ranger      & 1067              & mmce, nf, ias                      & Yes               \\
    iaml\_super       & 1489              & mmce, nf, ias                      & Yes               \\
    iaml\_super       & 1067              & mmce, nf, ias                      & Yes               \\
    iaml\_xgboost     & 40981             & mmce, nf, ias                      & Yes               \\
    iaml\_xgboost     & 1489              & mmce, nf, ias                      & Yes               \\
    iaml\_xgboost     & 40981             & mmce, nf, ias, rammodel            & Yes               \\
    iaml\_xgboost     & 1489              & mmce, nf, ias, rammodel            & Yes               \\
    lcbench           & 167152            & val\_accuracy, val\_cross\_entropy & Yes               \\
    lcbench           & 167185            & val\_accuracy, val\_cross\_entropy & Yes               \\
    lcbench           & 189873            & val\_accuracy, val\_cross\_entropy & Yes               \\
    rbv2\_ranger      & 6                 & acc, memory                        & Yes               \\
    rbv2\_ranger      & 40979             & acc, memory                        & Yes               \\
    rbv2\_ranger      & 375               & acc, memory                        & Yes               \\
    rbv2\_rpart       & 41163             & acc, memory                        & Yes               \\
    rbv2\_rpart       & 1476              & acc, memory                        & Yes               \\
    rbv2\_rpart       & 40499             & acc, memory                        & Yes               \\
    rbv2\_super       & 1457              & acc, memory                        & No                \\
    rbv2\_super       & 6                 & acc, memory                        & No                \\
    rbv2\_super       & 1053              & acc, memory                        & No                \\
    rbv2\_xgboost     & 28                & acc, memory                        & No                \\
    rbv2\_xgboost     & 182               & acc, memory                        & No                \\
    rbv2\_xgboost     & 12                & acc, memory                        & No                \\ \bottomrule
    \end{tabular}
    \caption{YAHPO-MO benchmark problems}
    \label{tab:apx-hpo-problems}
\end{table*}

\subsection{Noise Augmented ZDT}\label{apx:epx-zdt}
\begin{figure*}
    \centering
    \begin{subfigure}[t]{.2\linewidth}
        \includegraphics[width=\linewidth]{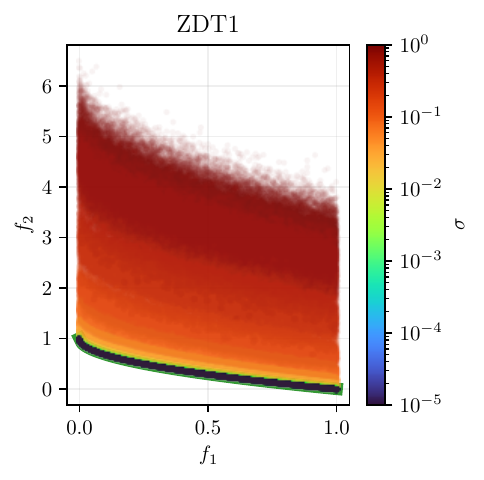}
    \end{subfigure}\hfill
    \begin{subfigure}[t]{.2\linewidth}
        \includegraphics[width=\linewidth]{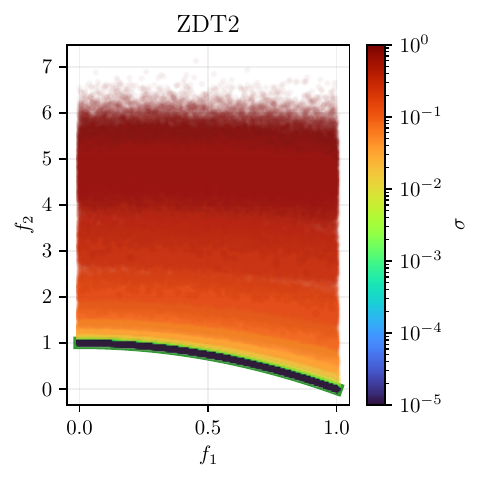}
    \end{subfigure}\hfill
    \begin{subfigure}[t]{.2\linewidth}
        \includegraphics[width=\linewidth]{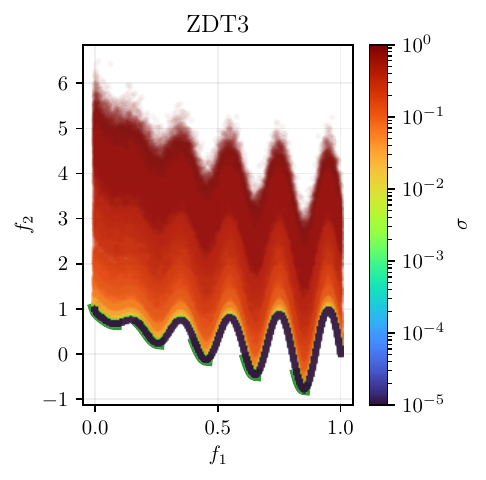}
    \end{subfigure}\hfill
    \begin{subfigure}[t]{.2\linewidth}
        \includegraphics[width=\linewidth]{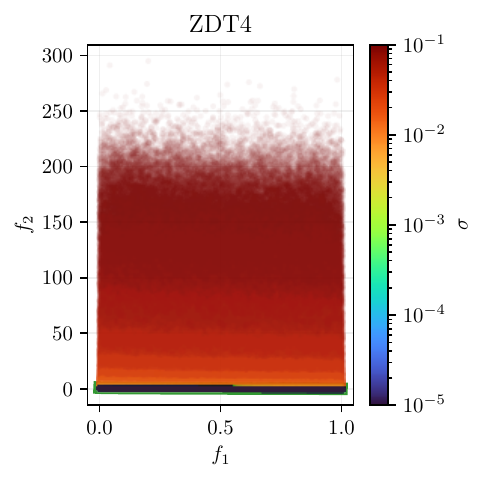}
    \end{subfigure}\hfill
    \begin{subfigure}[t]{.2\linewidth}
        \includegraphics[width=\linewidth]{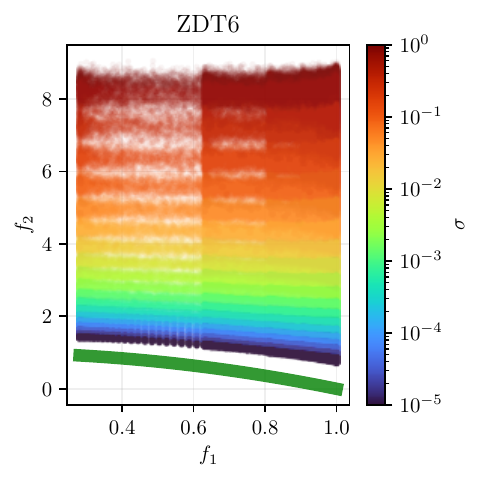}
    \end{subfigure}
    \caption{Scatter plots of the noise-augmented ZDT benchmarks, evaluated at the deterministic optimum, shown for several noise levels $\sigma$ (standard deviation of the input perturbation).}
    \label{fig:apx-zdt-noise}
\end{figure*}

We consider five out of six of the ZDT benchmark problems~\cite{ZDT}, excluding ZDT5, as the decision variables are binary; thus, our noise augmentation procedure does not apply. For each problem, we set the dimensionality of the input space to $N=30$. By adding noise to the input, we constructed optimization problems of the following form:
\begin{align*}
    \min_{\mathbf{x}}\quad &f(z_1,\dots,z_N),\\
    &z_i \sim \mathcal{N}_{[a_i,b_i]}(x_i,\sigma^2),
\end{align*}
where $a_i,b_i$ are the lower- and upper-bounds of the decision variable $x_i$, and $\mathcal{N}_{[a_i,b_i]}(x_i,\sigma^2)$ denotes the truncated normal distribution with mean $x_i$ and variance $\sigma^2$. In Figure~\ref{fig:apx-zdt-noise}, we show the objective values of the noise-augmented problems, evaluated at the deterministic optimum, for several values of $\sigma$.
Based on these plots, we chose to let $\sigma=0.1$ for all five problems, which was large enough to ensure that the objective values do not overlap with the deterministic Pareto front. Increasing $\sigma$ beyond a certain threshold would only alter the degree of the advantage of our method, not the mechanism.

For the implementation, we modified pymoo's \cite{pymoo} implementation of NSGA-II \cite{nsga2}. We use the default values for the crossover and mutation rates.
We chose a sample size of $n=64$, and $n_R=n_S=8$, to have sufficient samples to compute center-outward ranks and signs meaningfully.
We set the population size to $20$, and the number of generations to $200$ (or $200n$ for the single evaluation method). We varied the population size between $10$ and $100$, but found no significant difference in the respective conference rates. The number of generations was chosen to be approximately four times larger than what the fastest converging algorithm required to converge. This was done to show the convergence of the other, slower methods as well.
Each replication is re-seeded with a deterministic base-value-plus-index rule, yielding independent random streams.

\begin{figure*}
    \centering
    \includegraphics[width=\linewidth]{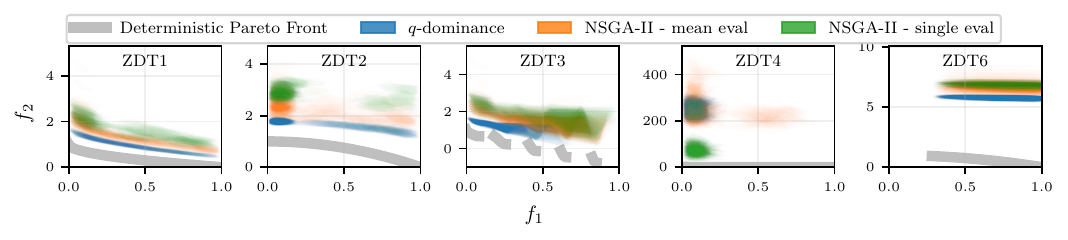}
    \caption{The $q\approx0.55$ quantile regions of the final Pareto sets.}
    \label{fig:apx-zdt-paretocontours}
\end{figure*}

In addition to the average $\Delta HV$ curves shown in Figure~\ref{fig:zdt-averageHV}, we also show, in Figure~\ref{fig:apx-zdt-paretocontours}, the $q\approx0.55$ center quantile regions of the final Pareto sets, over all $20$ replications, obtained after the complete evaluation budget of each method was used.

\end{document}